\documentclass[runningheads]{llncs}

\newcommand{\lepart}{\mbox{$\triangleleft$}}
\newcommand{\leqpart}{\mbox{$\trianglelefteq$}}

\newcommand{\complex}[1]{\mbox{$\mathcal{O}(#1)$}}
\newcommand{\curveMM}{\mbox{$-\bullet-$}}
\newcommand{\curveMMUn}{\mbox{$-\blacktriangle-$}}
\newcommand{\curveSM}{\mbox{$-+-$}}
\newcommand{\curveSMDeux}{\mbox{$-\Box-$}}
\newcommand{\curveSMCinq}{\mbox{\rule{4mm}{1mm}}}

\usepackage{epic,ecltree}

\usepackage{amsfonts,amssymb,txfonts}
\usepackage{epsfig,subfigure}
 
\begin{document}
\pagestyle{headings}
\mainmatter

\title{Hierarchy Construction Schemes within  the Scale Set Framework}
\titlerunning{Scale Set representation for image segmentation }

\author{Jean-Hugues PRUVOT \and Luc BRUN}
\authorrunning{Jean-Hugues PRUVOT \and Luc BRUN}

\institute{GREYC Laboratory, Image Team\\
CNRS UMR 6072\\
6, Boulevard Mar\'{e}chal Juin\\
14050 CAEN cedex FRANCE\\
\email{jhpruvot@greyc.ensicaen.fr}}

\maketitle

\begin{abstract}

  Segmentation algorithms based on an energy minimisation framework
  often depend on a scale parameter which balances a fit to data and a
  regularising term. Irregular pyramids are defined as a stack of
  graphs successively reduced. Within this framework, the scale is
  often defined implicitly as the height in the pyramid.  However,
  each level of an irregular pyramid can not usually be readily
  associated to the global optimum of an energy or a global criterion
  on the base level graph. This last drawback is addressed by the
  scale set framework designed by Guigues. The methods designed by
  this author allow to build a hierarchy and to design cuts within
  this hierarchy which globally minimise an energy.  This paper
  studies the influence of the construction scheme of the initial
  hierarchy on the resulting optimal cuts. We propose one sequential
  and one parallel method with two variations within both. Our
  sequential methods provide partitions near an energy lower bound
  defined in this paper. Parallel methods require less execution times
  than the sequential method of Guigues even on sequential machines.
  
\end{abstract}
\section{Introduction}
\label{sec:Introduction}

Despite much efforts and significant progresses in recent years, image
segmentation remains a notoriously challenging computer vision
problem. It's usually a preliminary step towards image interpretation
and plays a major role in many applications.

The use of an energy minimisation scheme within the region based
segmentation framework allows to define criteria which should be
globally optimised over a partition. Several types of methods such as
the Level set~\cite{lecellier-06}, the Bayesian~\cite{geman-84}, the
minimum description length~\cite{leclerc-89} and the minimal
cut~\cite{boykov-04} frameworks are based on this approach. Within
these frameworks the energy of a partition $P$ is usually defined as
$E_\lambda(P)=D(P)+\lambda C(P)$ where $D$ and $C$ denote respectively the fit to
data and the regularising term. The energy $E_\lambda(P)$ corresponds to the
Lagrangian of the constraint problem: minimise $D(P)$ subject to
$C(P)\leq \epsilon$.  Where $\epsilon$ is a function of $\lambda$. Under large assumptions,
minimising $E_\lambda(P)$ is also equivalent to the dual problem: minimise
$C(P)$ subject to $D(P)\leq \epsilon'$, where $\epsilon'$ is also a function of $\lambda$.
Therefore $\lambda$ may be interpreted as the amount of freedom allowed to
minimise $D$ ($D(P)\leq \epsilon'$) while keeping $C$ as low as possible. Since
$\epsilon'$ is a growing function of $\lambda$, as $\lambda$ is growing, the constraint
on $D$ is more and more relaxed while the importance of the term $C$
is getting more and more important.  This parameter $\lambda$ may thus be
interpreted as a \emph{scale parameter} which represents the relative
weighting between the two energy terms.

In many approaches the parameter $\lambda$ is fixed experimentally and a
minimisation algorithm determines for a value of $\lambda$ a locally optimal
partition from the set $\mathbb{P}$ of all the possible partitions on
image $I$. A sequence of $\lambda$ may also be defined a priori in order to
compute the optimal partition on each sampled value of
$\lambda$~\cite{guigues03}.

The scale set framework proposed by Guigues~\cite{guigues03} is based
on a different approach. Instead of performing the minimisation scheme
on the whole set $\mathbb{P}$ of possible partitions of an image $I$,
Guigues proposes to restrict the search on a hierarchy $H$. The
advantages of this approach are twofold: firstly as shown by Guigues the
globally optimal partition on $H$ may be found efficiently while the
search on the whole set $\mathbb{P}$ of partitions only provides local
minima.  Secondly, Guigues shown that if the energy satisfies some
basic properties, the whole set of solutions on $H$ when $\lambda$ describes
$\mathbb{R}+$ corresponds to a sequence of increasing cuts within the
hierarchy $H$ hereby providing a contiguous representation of the
solutions for the parameter $\lambda$. A method to build the hierarchy $H$
has been proposed by Guigues. Since the research space used by Guigues
is restricted to the initial hierarchy $H$ the construction scheme of
this hierarchy is of crucial importance for the optimal partitions
\emph{within $H$} built in the second step.

This paper explores different heuristics to build the initial
hierarchy. These heuristics represent different compromises between
the energy of the final partitions and the execution times. We first
present in Section~\ref{sec:Guigues} the scale set framework. The
different heuristics are then presented in
Section~\ref{sec:MultiFusion}. These heuristics are evaluated and
compared to the method of Guigues in Section~\ref{sec:Experiments}.


\section{The Scale Set framework}

\label{sec:Guigues}

Given an image $I$ and two partitions $P$ and $Q$ on $I$, we will say
that $P$ is \emph{finer} than $Q$ (or $Q$ is coarser then $P$) iff $Q$
may be deduced from $P$ by merging operations. This relationship is
denoted by $P\leqpart Q$. Let us now consider a theoretic segmentation
algorithm $P_\lambda$ parametrised by $\lambda$. We will say that $P$ is an
\emph{unbiased multi-scale segmentation} algorithm iff for any couple
$(\lambda_1,\lambda_2)$ such that $\lambda_1\leq \lambda_2$, and any image $I$,
$P_{\lambda_1}(I)\leqpart P_{\lambda_2}(I)$.  If $P_\lambda$ is an unbiased multi-scale
segmentation algorithm, $P_\lambda(I)$ increases according to $\lambda$ and the
set $H=\bigcup_{\lambda\in \mathbb{R}+}P_\lambda(I)$ defines a hierarchy as an union of
nested partitions. Note that the set $\mathbb{P}$ of partitions on $I$
being finite, $H$ must be also finite.

Unbiased multi-scale segmentation algorithms follow a well known
causal principal: increasing the scale of observation should not
create new information.  In other words any phenomenon observed at one
scale should be caused by objects defined at finer scales. In our
framework, increasing the scale should not create new contours.

The family of energies considered by Guigues corresponds to the set
of Affine Separable Energies (ASE) which can be written for any
partition $P$ of $I$ in $n$ regions $\{R_1,\dots,R_n\}$ as:
\[
E(P)=D(P)+\lambda C(P)=
\sum_{i=1}^nD(R_i)+\lambda\sum_{i=1}^nC(R_i)=
\sum_{i=1}^nD(R_i)+\lambda C(R_i)
\]

Let us consider a hierarchy $H$ and the sequence
$(C^*_\lambda(H))_{\lambda\in \mathbb{R}+}$ of optimal cuts within
$H$. The approach of Guigues is based on the following result: If
$E_\lambda(P)$ is an ASE and if $C_\lambda(P)$ is decreasing within
$\mathbb{P}$:
\[
\forall (P,Q)\in \mathbb{P}\quad P\lepart Q\Rightarrow C(P)> C(Q)
\]
then the sequence $(C^*_\lambda(H))_{\lambda\in \mathbb{R}+}$ is an unbiased
multi-scale segmentation. The union of all $(C^*_\lambda(H))_{\lambda\in
  \mathbb{R}+}$ defines thus a new hierarchy within $H$. The tree
corresponding to the hierarchical structure of $\bigcup_{\lambda\in \mathbb{R}+}
C^*_\lambda(H)$ may be deduced from $H$ by merging with their fathers all
the nodes which do not belong to any optimal cuts. Note that an
equivalent result may be obtained if no condition is imposed to $C$
but if $D$ is increasing according to $\lambda$.

The restriction by Guigues of the research space to a hierarchy may
thus be justified by the fact that the set of partitions produced by
any unbiased multi-scale segmentation algorithm describes a hierarchy.
Conversely, given a hierarchy $H$, if the energy $E_\lambda$ is an ASE with
a decreasing term $C$ the sequence of optimal cuts of $H$ according to
$E_\lambda$: $(C^*_\lambda(H))_{\lambda\in\mathbb{R}+}$ is an unbiased multi-scale
segmentation algorithm.

Given a partition $P\in\mathbb{P}$, the decrease of $C$ may be
equivalently expressed as a sub-additivity relationship:
\begin{equation}
\label{eq:subAdd}
  \forall (R,R')\in P~|~ R\mbox{ is adjacent to }R' \quad C(R\cup R')< C(R)+C(R')
\end{equation}

Note that the sub-additivity of the regularising term $C$ in common is
many applications. For example, if $C$ is proportional to some
quantity summed up along contours, $C$ is sub-additive due to the
removal of the common boundaries between the two merged regions.
Moreover, the term $C$ may be interpreted within the Minimum
Description Length framework~\cite{leclerc-89} as the amount of
information required to encode a partition. Therefore, one can expect
$C$ to decrease when the partition gets coarser.

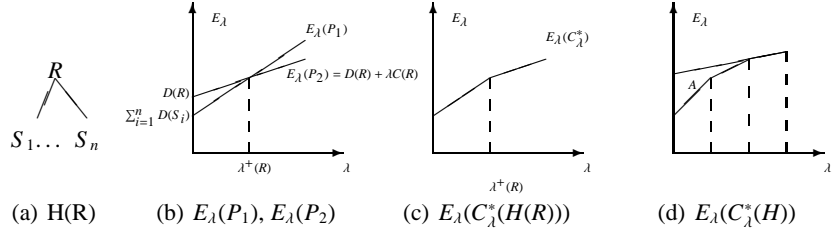
\begin{figure}[t]
  \unitlength .5mm
  \centering
\mbox{ }\hfill
  \subfigure[H(R)]{
    \begin{minipage}[b]{0.1\linewidth}
    \begin{bundle}{$R$}
      \chunk{$S_1$\dots}  \chunk{$S_n$}
    \end{bundle}
    \vspace{5mm}
  \end{minipage}}
\hfill
\subfigure[$E_\lambda(P_1)$, $E_\lambda(P_2)$]{
\begin{picture}(50,50)
    \tiny
    \put(10,10){\vector(1,0){40}}
    \put(10,10){\vector(0,1){40}}
    \drawline(10,20)(40,40)
    \drawline(10,25)(40,35)
    \put(40,42){\centering  $E_\lambda(P_1)$}
    \put(-8,19){$\sum_{i=1}^n D(S_i)$}
    \put(2,25){$D(R)$}
    \put(35,30){\centering $E_\lambda(P_2)=D(R)+\lambda C(R)$}
    \dashline{3}(25,30)(25,10)
    \put(22,5){\centering $\lambda^+(R)$}
    \put(50,5){$\lambda$}
    \put(15,45){$E_\lambda$}
  \end{picture}}
  \hfill
  \subfigure[$E_\lambda(C^*_\lambda(H(R)))$]{
    \begin{picture}(50,50)
    \tiny
    \put(10,10){\vector(1,0){40}}
    \put(10,10){\vector(0,1){40}}
    \drawline(10,20)(25,30)(40,35)
    \dashline{3}(25,30)(25,10)
    \put(25,0){\centering $\lambda^+(R)$}
    \put(50,5){$\lambda$}
    \put(15,45){$E_\lambda$}
    \put(40,40){$E_\lambda(C^*_\lambda)$}
  \end{picture}}
\hfill
  \subfigure[$E_\lambda(C^*_\lambda(H))$]{
    \begin{picture}(50,60)
    \tiny
    \put(10,10){\vector(1,0){40}}
    \put(10,10){\vector(0,1){40}}
    \drawline(10,20)(20,30)(30,35)(40,37)
    \drawline(10,31)(40,37)
    \put(14,27){$A$}
    \dashline{3}(20,30)(20,10)
    \dashline{3}(30,35)(30,10)
    \dashline{3}(40,37)(40,10)
    \put(15,45){$E_\lambda$}
    \put(50,5){$\lambda$}
  \end{picture}}
\hfill\mbox{ }
\caption{(a) a node $R$ of the hierarchy whose sons $\{S_1,\dots,S_n\}$
  correspond to initial regions. (b) the energies of the partitions
  associated to $R$ and $\{ S_1,\dots,S_n\}$ plotted as functions of
  $\lambda$. (c) the energy of the optimal cuts within $H(R)$ (a).  (d) an
  example of concave piecewise linear function encoding the energy of
  the optimal cuts within a global hierarchy $H$.}
  \label{fig:lambdaPlus}
\end{figure}

Given a hierarchy $H$, the sequence of optimal cuts $C^*_\lambda(H)$ within
$H$ has to be computed. Let us consider one region $R$ at the second
level of the hierarchy (computed from the base) and its set of sons
$S_1,\dots,S_n$. Let us additionally consider the tree $H(R)$ rooted
at $R$ within $H$ (Fig.~\ref{fig:lambdaPlus}(a)). Since $R$ is a level
$2$ node, the hierarchy $H(R)$ allows only two cuts: one encoding the
partition $P_1$ made of the sons of $R$ whose energy is equal to
$E_\lambda(P_1)=\sum_{i=1}^nD(S_i)+\lambda\sum_{i=1}^nC(S_i)$ and one encoding the
partition $P_2$ reduced to the single region $R$. The energy of $P_2$
is equal to $E_\lambda(P_2)=D(R)+\lambda C(R)$. Due to the sub additivity of $C$
we have $\sum_{i=1}^nC(R_i)>C(R)$.  Therefore, using the linear
expression of $E_\lambda(P_1)$ and $E_\lambda(P_2)$ in $\lambda$, if $\sum_{i=1}^nD(S_i) <
D(R)$ the line $E_\lambda(P_1)=\sum_{i=1}^nD(S_i)+\lambda\sum_{i=1}^nC(R_i)$ is below
the line $E_\lambda(P_2)=D(R)+\lambda C(R)$ until a value $\lambda^+(R)$ of $\lambda$ for
which the two lines cross(Fig.~\ref{fig:lambdaPlus}(b)). If
$\sum_{i=1}^nD(S_i) \geq D(R)$, $E_\lambda(P_2)$ is always greater or equal to
$E_\lambda(P_1)$ in which case we set $\lambda^+(R)$ to $0$.  Therefore, in both
cases the partition $P_1$ is associated to a lower energy than $P_2$
for $\lambda=0$ until $\lambda=\lambda^+(R)$. Above this value the partition $P_2$ is
associated to the lowest energy. In terms of optimal cuts, $P_1$
corresponds to the optimal cut of $H(R)$ until $\lambda^+(R)$ and $P_2$ is
the optimal cut above this value(Fig.~\ref{fig:lambdaPlus}(c)). The
value $\lambda^+(R)$ is called the \emph{scale of appearance} of the region
$R$.

Guigues shown that the above process may be generalised to the whole
tree. Each node of $H$ is then valuated by a scale of appearance.
Some of the nodes of $H$ may get a greater scale of appearance than
their father. Such nodes do not belong to any optimal cut and are
removed from $H$ during a cleaning step which merges them with their
fathers. Each node $R$ of the resulting hierarchy belongs to an
optimal cut from $\lambda=\lambda^+(R)$ until the scale of appearance of its
father $\lambda^+(\mathcal{F}(R))$, where $\mathcal{F}(R)$ denotes the
father of $R$ in $H$.  The value $\lambda^+(R)$ may be set for each node of
the tree using a bottom-up process. The optimal cut $C^*_\lambda(H)$ for a
given value of $\lambda$ may then be determined using a top-down process
which selects in each branch of the tree the first node with a scale
of appearance lower than $\lambda$. The set of selected nodes constitutes a
cut of $H$ which is optimal by construction according to $E_\lambda$. The
function $E_\lambda(C^*_\lambda(H))$ corresponds to a concave piecewise linear
function whose each linear interval corresponds to the energy of an
optimal cut within $H$ (Fig.~\ref{fig:lambdaPlus}(d)). 

Given a hierarchy $H$ and the function $E_\lambda(C^*_\lambda(H))$ encoding the
energy of the sequence of optimal cuts, the optimality of $H$ may be
measured as the area under the curve $E_\lambda(C^*_\lambda(H))$ for a given range
of scales or as the area of the surface $A$
(Fig.~\ref{fig:lambdaPlus}(d)) between $E_\lambda(C^*_\lambda(H))$ and the energy
of the coarsest cut $E_\lambda(P_{max})$. Where $P_{max}$ denote the
partition composed of a single region encoding the whole image. We
propose in Section~\ref{sec:Experiments} an alternative measure of the
quality of a hierarchy which allows to reduce the influence of the
initial image.

Guigues proposed to build a hierarchy $H$ by using an initial
partition $P_0$ and a strategy called the \emph{scale climbing}. This
strategy merges at each step the two adjacent regions $R$ and $R'$
such that:
\begin{equation}
  \lambda^+(R\cup R')=\frac{D(R\cup R')-D(R)-D(R')}{C(R)+C(R')-C(R\cup 
R')}=\min_{(R_1,R_2)\in P^2,R_1\sim R_2}\frac{D(R_1\cup R_2)-D(R_1)-D(R_2)}{C(R_1)+C(R_2)-C(R_1\cup 
R_2)}\label{eq:scaleClimbing}
\end{equation}
where $P$ denotes the current partition and $R_1\sim R_2$ indicates that
$R_1$ and $R_2$ are adjacent in $P$.

This process merges thus at each step the two regions whose union
would appear at the lowest scale.  Such a construction scheme is
coherent with the further processes applied on the hierarchy. However,
there is no evidence that the resulting hierarchy may be optimal
according to any of the previously mentioned criteria.  We indeed show
in the next section that other construction schemes of a hierarchy may
lead to lower energies.


\section{Construction of the initial hierarchy}
\label{sec:MultiFusion}

Many energies have been designed in order to encode different types of
homogeneity criteria (piecewise
constant~\cite{leclerc-89,mumfordShah89}, linear or
Polynomial~\cite{leclerc-89} variations,\dots). This paper being
devoted to the construction schemes of the hierarchy, we restrict our
topic to the piecewise constant model described by
Leclerc~\cite{leclerc-89} and Mumford and Shah~\cite{mumfordShah89}.
The energy of this model may be written as:
\begin{equation}
  E_\lambda(P)=D(P)+\lambda C(P)=\sum_{i=1}^nS\!E(R_i)+\lambda|\delta(R_i)|\label{eq:mumford}
\end{equation}
where $P=\{R_1,\dots,R_n\}$ represents the partition of the image,
$S\!E(R_i)=\sum_{p\in R}\|c_p -\mu_R\|^2$ is the squared error of region $R_i$
and $|\delta(R_i)|$ is the total length of its boundaries.

Within the Minimum Description Length framework, $SE(R_i)$ may be
understood as the amount of information required to encode the
deviation of the data against the model, while $|\delta(R_i)|$ is
proportional to the amount of information required to encode the shape
of the model. Within the statistical framework, the squared error may
also be understood as the log of the probability that the region
satisfies the model (i.e. is constant) using a Gaussian assumption
while $|\delta(R_i)|$ is a regularising term.

Our approach follows the scale climbing strategy proposed by Guigues
(equation~\ref{eq:scaleClimbing}).  Given a set $W$ of regions within
a partition $P$ we thus consider the scale of appearance of the region
$R$ defined as the union of the regions in $W$. The heuristics below
use this basic approach but differ on the sets $W$ which are
considered and on the ordering of the merge operations.

\subsection{Sequential Merging}
\label{subsec:sequential}
Given a current partition $P$, let us consider for each region $R$ of
$P$, its set $V(R)$ defined as $\{R\} $ union its set of neighbours and
the set $\mathcal{P}^*(V(R))$ of all possible subsets of $V(R)$
including $R$.  Each subset $W\in \mathcal{P}^*(V(R))$ encodes a
possible merging of the region $R$ with at least one of its neighbour.
Let us denote by $R^W=\bigcup_{R'\in W} R'$ the region formed by the union of
the regions in $W$. Note that the region $R^W$ is connected since $R$
belongs to $W$ and all the regions of $W$ are adjacent to $R$. Let us
additionally consider the two partitions of $R^W$: $P_{R^W}=\{R^W\}$ and
$P_W=W$. The energies associated to these partitions are respectively
equal to $E_\lambda(P_{R^W})=D(R^W)+\lambda C(R^W)$ and:
\[
E_\lambda(P_W)=D(W)+\lambda C(W)=\sum_{R'\in W} D(R')+\lambda\sum_{R'\in W} C(R') 
\]
where $D(W)$ and $C(W)$ denote respectively the fit to data and the
regularising terms of the partition $P_W$.

Since $C$ is sub additive (equation~\ref{eq:subAdd}) we have
$C(W)>C(R^W)$.  The energy $E_\lambda(P_W)$ is thus lower than
$E_\lambda(P_{R^W})$ until a value $\lambda^+(R^W)$ called the scale of appearance
of $R^W$ (Section~\ref{sec:Guigues}).  Using the scale climbing
principle, our sequential merging algorithm computes for each region
$R$ of the partition the minimal scale of appearance of a region
$R^W$:

$$
\lambda^+_{min}(R)=\arg min_{W\in \mathcal{P}^*(V(R))}\frac{D(R^W)- D(W)}{C(W)-C(R^W) }
$$
the set $W\in \mathcal{P}^*({V(R)})$ which realises the min is denoted
$W_{min}(R)$.

Given the quantities $\lambda^+_{min}(R)$ and $W_{min}(R)$, our sequential
algorithm iterates the following steps:
\begin{enumerate}
\item Let $P$ denotes the current partition initialised with an initial
  partition $P_0$,

\item\label{item:lambdaPlus} For each region $R$ of $P$ compute
  $\lambda^+_{min}(R)$ and $W_{min}(R)$

\item Compute $R_{min}=\arg min_{R\in P}\lambda^+_{min}(R)$ and merge all the
  regions of $W_{min}(R_{min})$. 

\item If more than one region remains go to step
  \ref{item:lambdaPlus},

\item Output the final hierarchy $H$ encoding the sequence of merge
  operations.

\end{enumerate}

This algorithm performs thus one merge operation at each step of the
algorithm.  Note that all the regions of $W_{min}(R_{min})$ are
adjacent to $R_{min}$. Therefore, within the irregular pyramid
framework, the merge operation may be encoded by a contraction kernel
of depth one composed of a single tree whose root is equal to
$R_{min}$. The computation of $\lambda^+_{min}(R)$ for each region $R$ of
the partition requires to traverse $\mathcal{P}^*(V(R))$ whose
cardinal is equal to $2^{|V(R)|-1}$. Therefore, if the partition is
encoded by a graph $G=(V,E)$, the complexity of each step of our
algorithm is bounded by $\complex{|V|2^k}$ where $|V|$ denotes the
number of vertices (i.e.  the number of regions) and $k$ represents
the maximal vertices's degree of $G$. The cardinal of $V$ is decreased
by $|W_{min}(R_{min})|-1$ at each iteration.  Since
$|W_{min}(R_{min})|$ is at least equal to $2$, the cardinal of $V$
decreases by at least $1$.  The computation of $\lambda^+_{min}(R)$ for each
region $R$ of the partition may induce important execution times when
the degree of the vertices of the graph is important. However,
experiments presented in Section~\ref{sec:Experiments} show that the
cardinal of the subsets $W\in \mathcal{P^*}(R)$ may be bounded without
altering significantly the energy of the optimal cuts.  Let us finally
note that this algorithm includes the scale climbing approach proposed
by Guigues.  Indeed, the merge operations studied by Guigues
(Section~\ref{sec:Guigues}) correspond to the subsets $W\in
\mathcal{P}^*(V(R))$ with $|W|=2$ which are considered by our
algorithm.

\subsection{Parallel Merge algorithm}

\label{subsec:parallel}

Our parallel merge algorithm is based on the notion of maximal
matching. A set of edges $M$ of a graph $G=(V,E)$ is called a maximal
matching if each vertex of $G$ is incident to at most one edge of $M$
and if $M$ is maximal according to this property. Moreover, we would
like to design a maximal matching $M$ such that the scale of
appearance of the regions produced by the contraction of $M$ is as
low as possible. Let us denote by $\iota(e)$, the two vertices incident to
$e$. Using the same approach as in Section~\ref{subsec:sequential} we
associate to each edge $e$ of the graph the scale of appearance
$\lambda^+(\iota(e))$ (equation~\ref{eq:scaleClimbing}) of the region $R^{\iota(e)}$
defined as the union of the regions encoded by the two vertices
incident to $e$.  Following, the same approach as
Haxhimusa~\cite{haxhimusa-03} we define our maximal matching as a
Maximal Independent Set on the set of edges of the graph. The
iterative process which builds the maximal independent set selects at
each step edges whose scale of appearance is locally minimal. This
process may be formulated thanks to two boolean variables $p$ and $q$
attached to each edge such that:
\begin{equation}  
\left\{\begin{array}[c]{lll}
  p_e^1&=&\lambda^+(e)=min_{e'\in \Gamma(e)} \{ \lambda^+(e')\} \\
  q_e^1&=&\bigwedge_{e'\in \Gamma(e)}\overline{p_{e'}^1}\\
\end{array}\right.
\mbox{ and }
\left\{\begin{array}[c]{lll}
  p_e^{k+1}&=&p_e^k\lor \left(q_e^k\land \lambda^+(e)=min_{e'\in \Gamma(e)~|~  q_{e'}^k} \{ \lambda^+(e')\}\right) \\
  q_e^{k+1}&=&\bigwedge_{e'\in \Gamma(e)}\overline{p_{e'}^{k+1}}\\
\end{array}\right.\label{eq:mis}
\end{equation}
where $\Gamma(e)$ denotes the neighbourhood of the edge $e$ and is defined
as $\Gamma(e) = \{e\} \cup \{e' \in E | \iota(e)\cap \iota(e')\neq\emptyset \}$.

This iterative process stops when no change occurs between two
iterations. If $n$ denotes the final iteration, the set of edges such
that $p_e^n$ is true defines a maximal matching~\cite{haxhimusa-03}
$M$ which encodes the set of edges to be contracted. Moreover, the set
of selected edges corresponds to local minima according to the scale
of appearance $\lambda^+(e)$.  Roughly speaking if $\lambda^+(e)$ is understood as
a merge score, one edge between two vertices will be marked
$(p_e^k=true)$ at iteration $k$, if among all the remaining possible
merge operations involving these two vertices, the one involving them
is the one with the best merge score. Note that the construction of a
maximal matching is only the first step of the method of Haxhimusa
which completes this maximal matching in order to get a decimation
ratio of order $2$. The restriction of our method to a maximal
matching allows to restrict the merge operations to edges which become
locally optimal at a given iteration. We thus favour the energy
criterion against the reduction factor.  As shown by
Bield~\cite{maxMatching04}, the reduction factor in terms of edges
induced by the use of a maximal matching is a least equal to
$2\frac{k-1}{2k-1}$ where $k$ is the maximal vertex's degree of the
graph. The edge's decimation ratio may thus be  very low
for graphs with important vertices's degrees.  Nevertheless,
experiments performed on 100 natural images of the Berkeley
database\footnote[1]{available at
  http://www.eecs.berkeley.edu/Research/Projects/CS/vision/bsds/}
shown that the mean vertex's decimation ratio between levels on this
database is equal to $1.73$ which is comparable to the $2.0$
decimation ratio obtained by Haxhimusa.

The local minima selected in equation~\ref{eq:mis} are computed on
decreasing sets along the iterations in order to complete the maximal
matching. We can thus consider that the detected minima are less and
less significants as the iterations progress. We thus propose an
alternative solution which consists in contracting at each step only
the edges selected at the first iteration ($p_e^1=true$). These edges
correspond to minima computed on the whole neighbourhood of each edge.
This method may be understood as a combination of the method proposed
by Haxhimusa~\cite{haxhimusa-03} and the stochastic decimation process
of Jolion~\cite{jolion-01} which consists in merging immediately
vertices corresponding to local minima.


\section{Experiments}
\label{sec:Experiments}

\begin{figure}[t] 
\centering 
 \begin{tabular}{ccccccccc}  

\begin{minipage}[b]{0.05\linewidth} 
  MM 
  \\ 
  \vspace{4mm} 
\end{minipage} 
&  \includegraphics[width=0.11\linewidth]{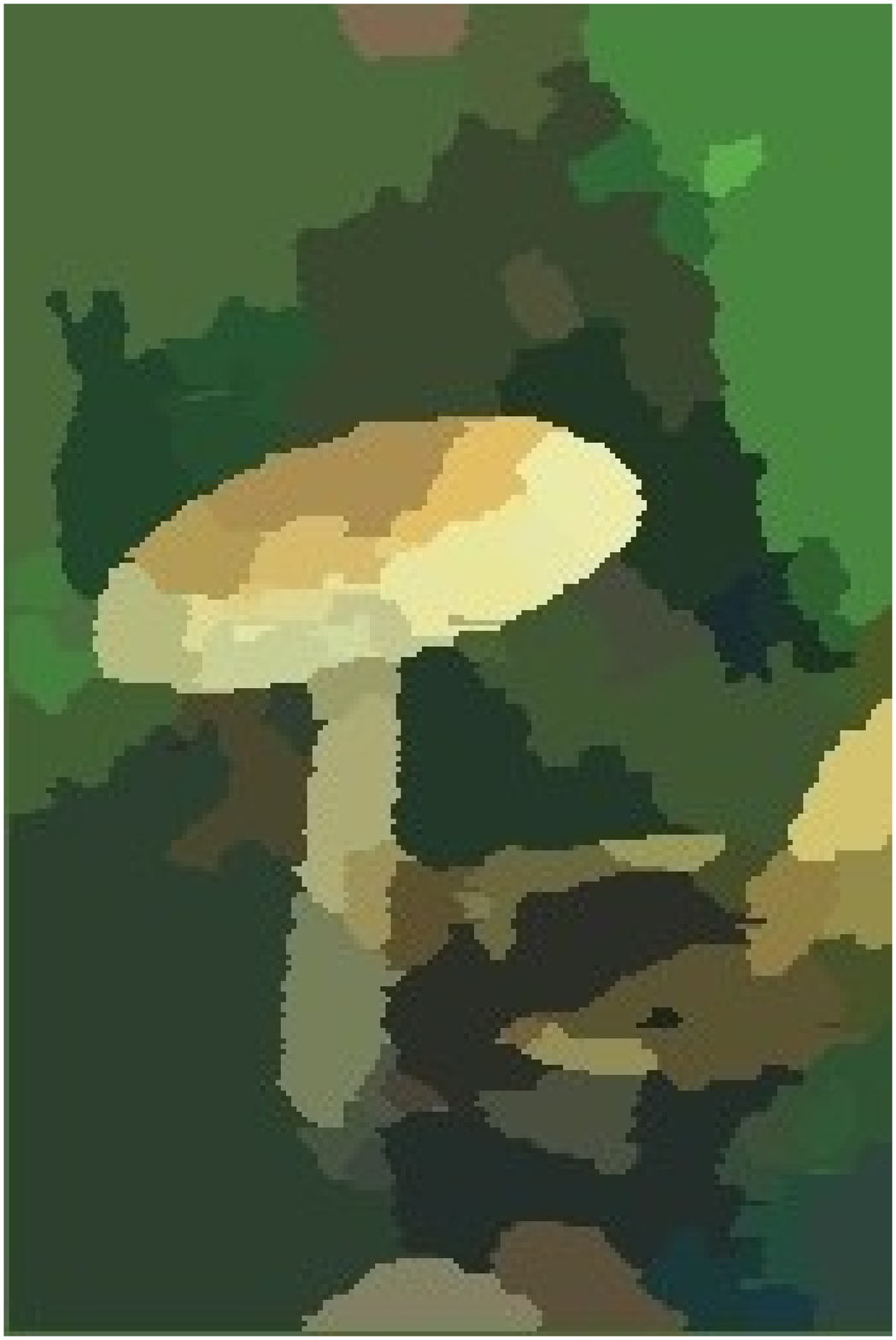} & \includegraphics[width=0.11\linewidth]{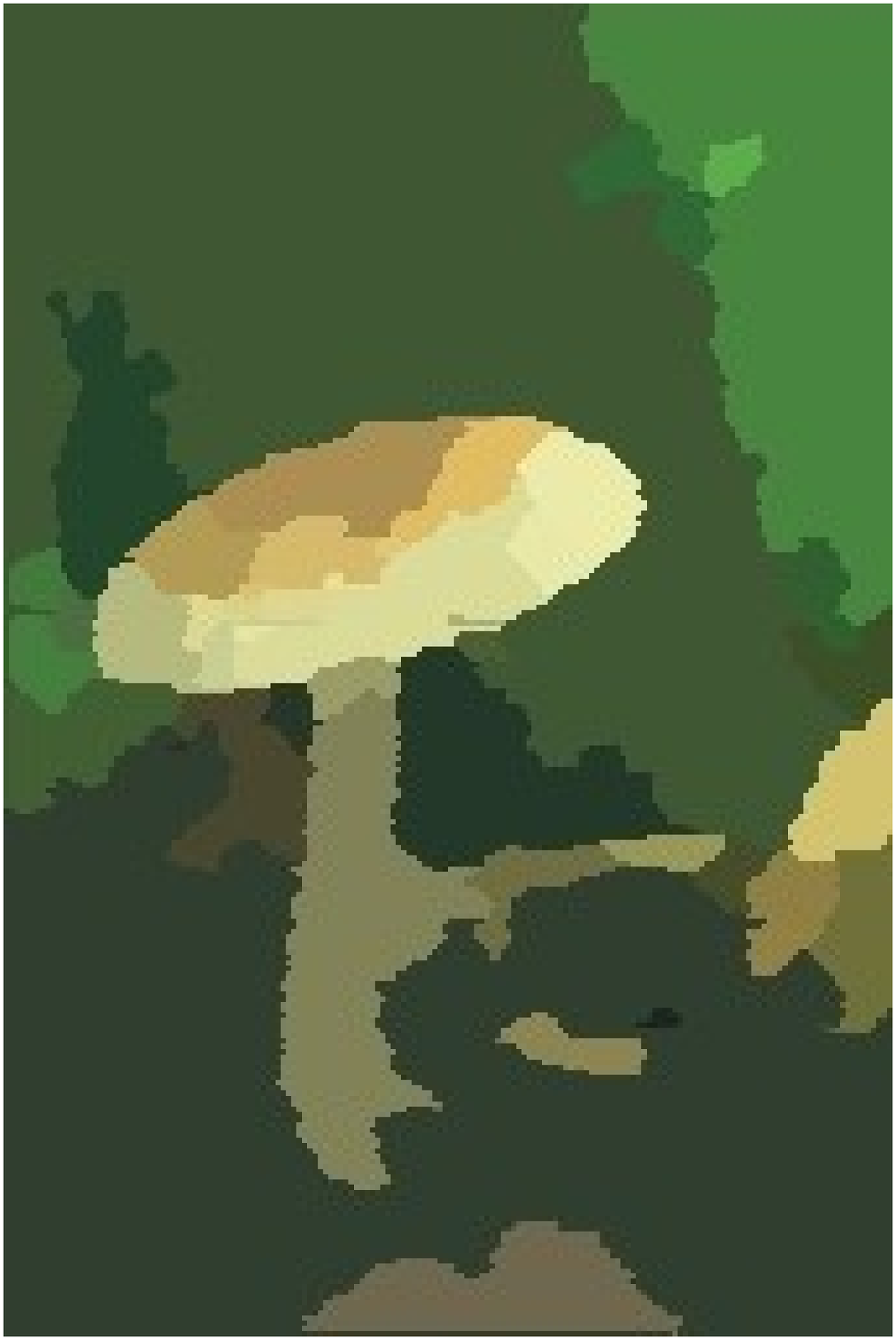} & \includegraphics[width=0.11\linewidth]{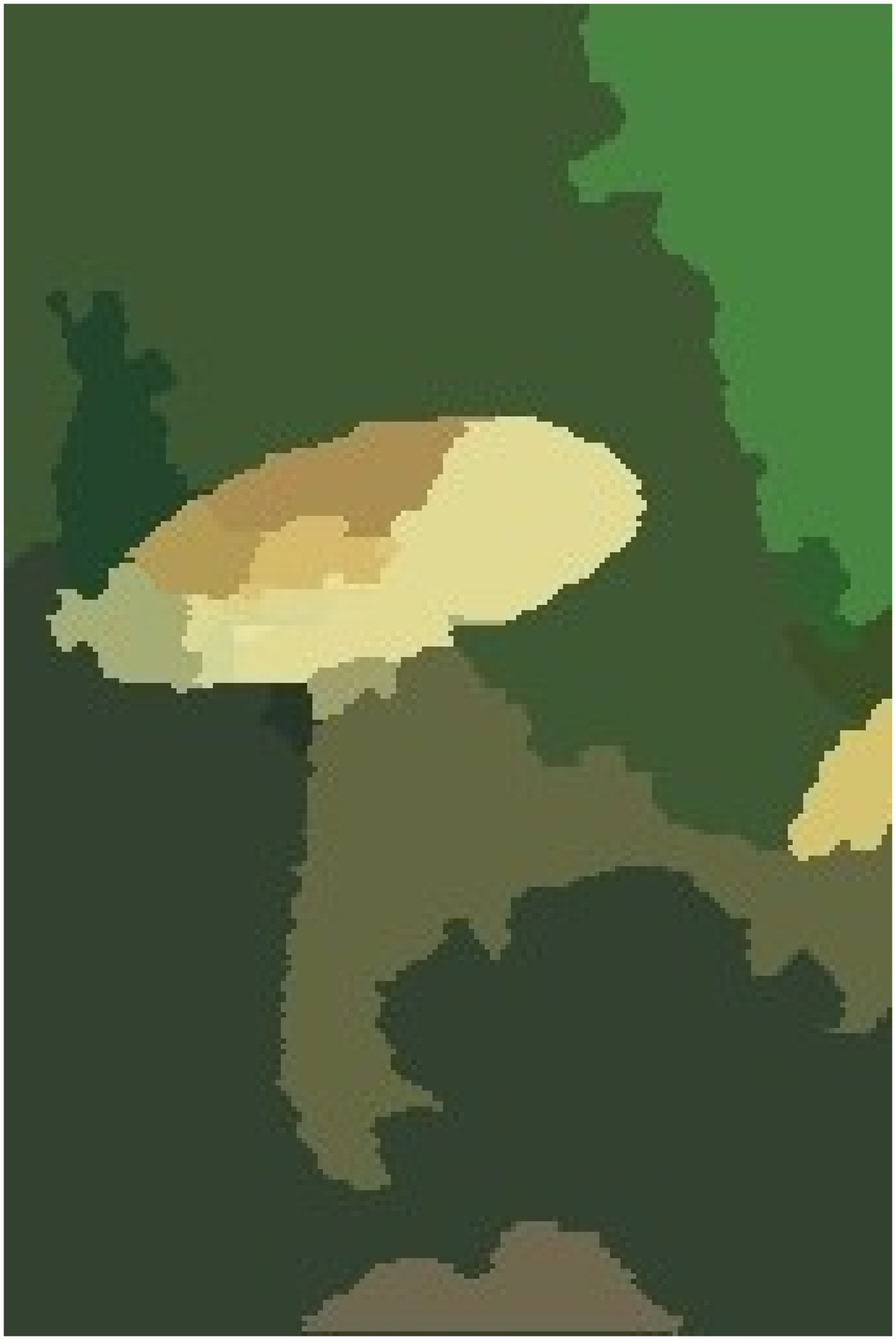}& \includegraphics[width=0.11\linewidth]{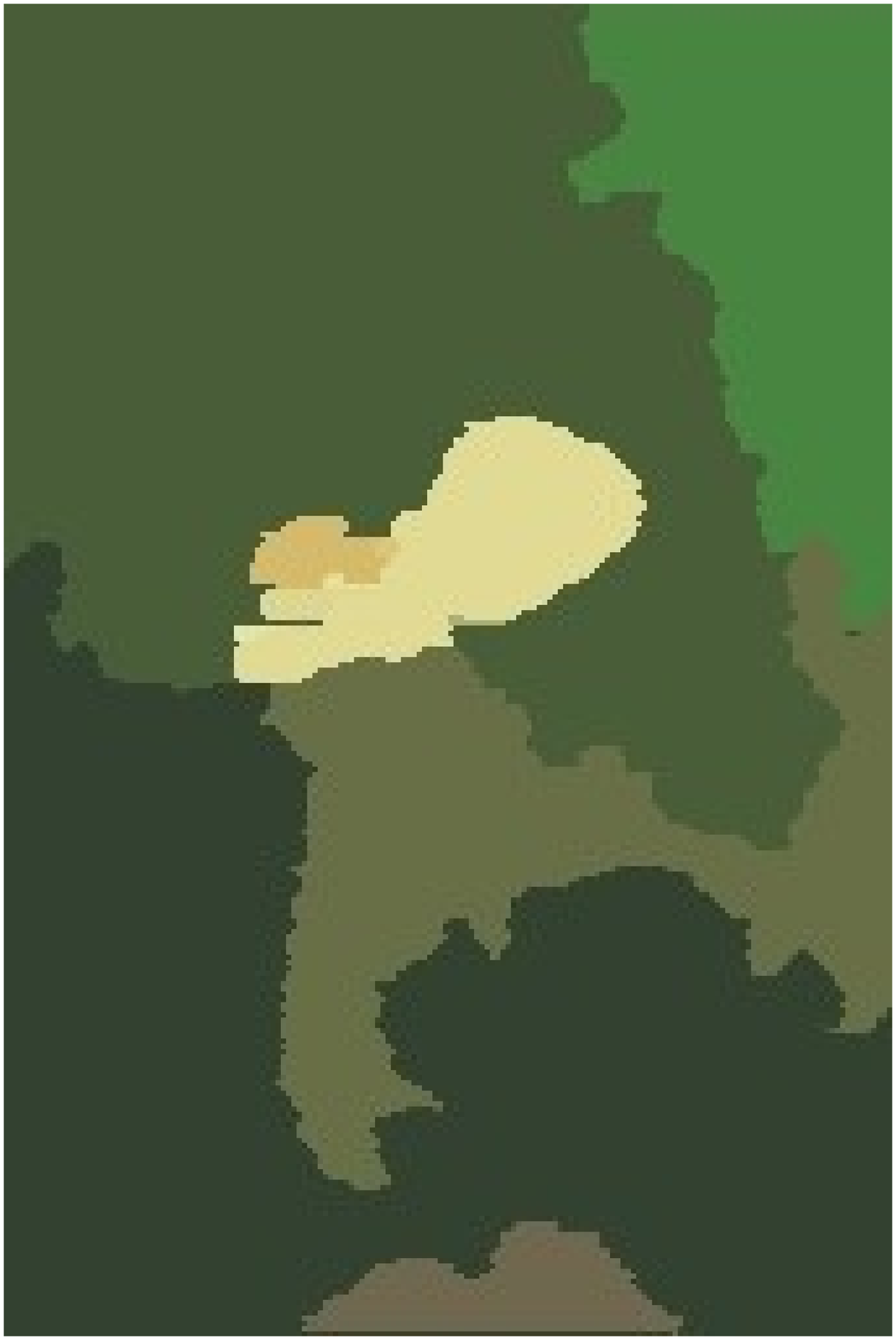} &\includegraphics[width=0.11\linewidth]{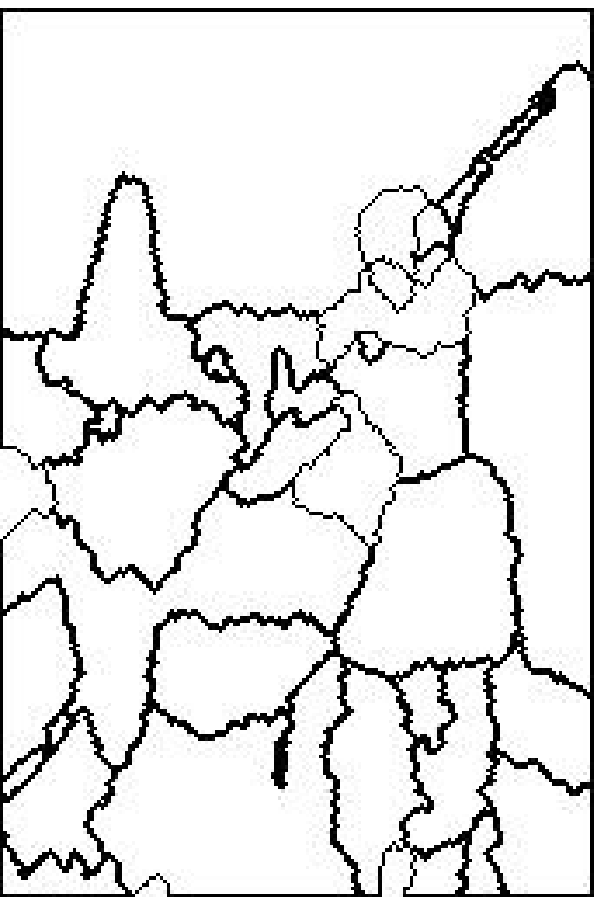} & \includegraphics[width=0.11\linewidth]{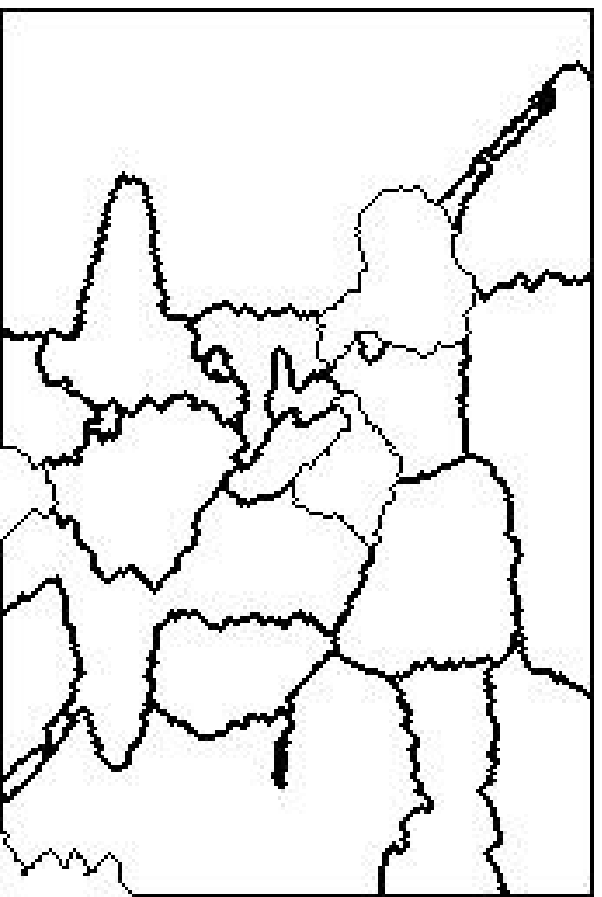} & \includegraphics[width=0.11\linewidth]{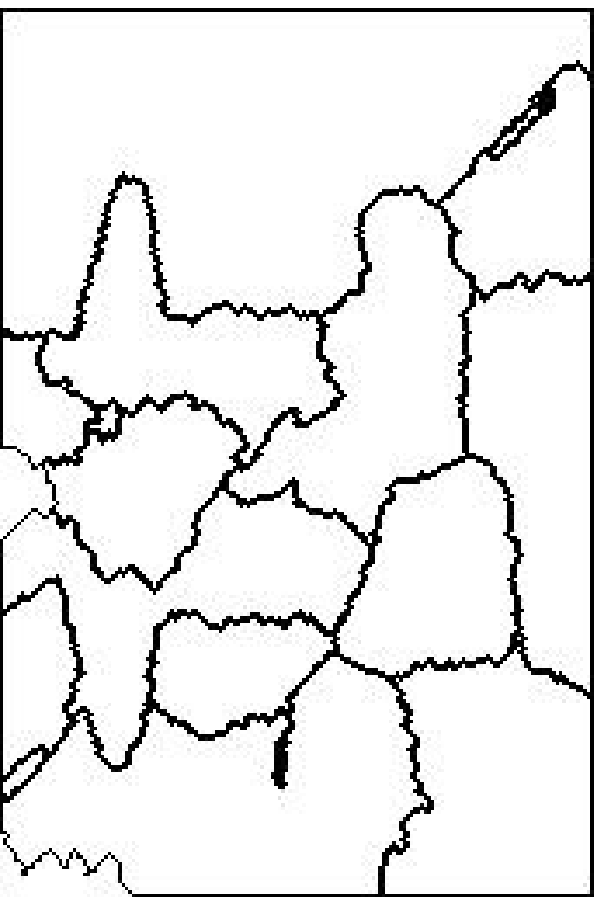}& \includegraphics[width=0.11\linewidth]{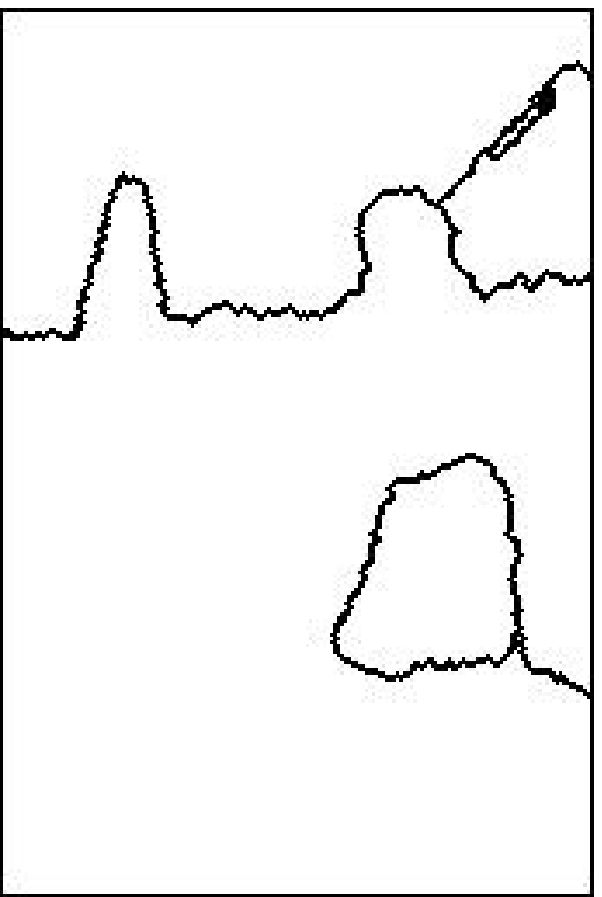}\\

\begin{minipage}[b]{0.05\linewidth} 
  $MM^1$ 
  \\ 
  \vspace{4mm} 
\end{minipage}  
&  \includegraphics[width=0.11\linewidth]{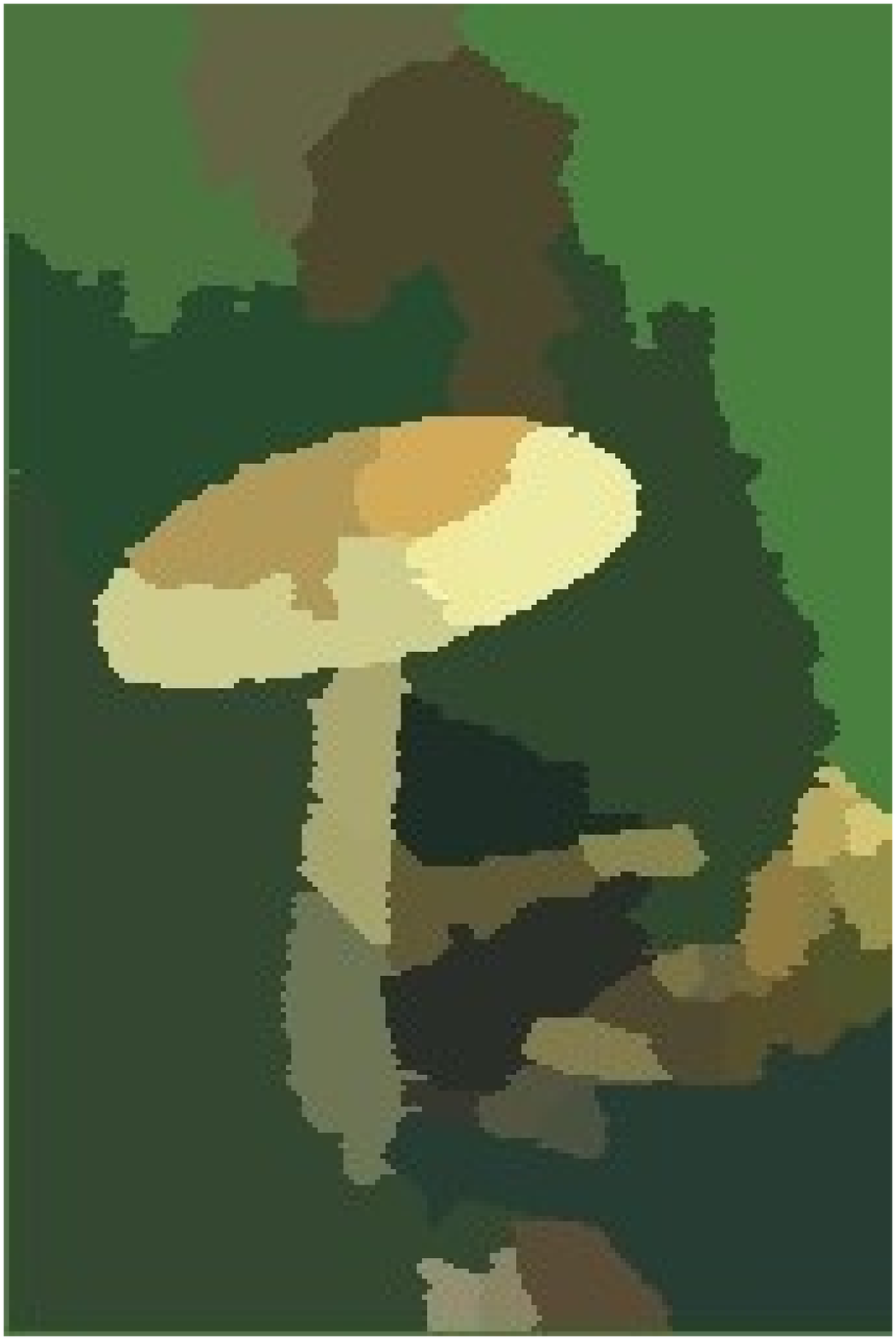} & \includegraphics[width=0.11\linewidth]{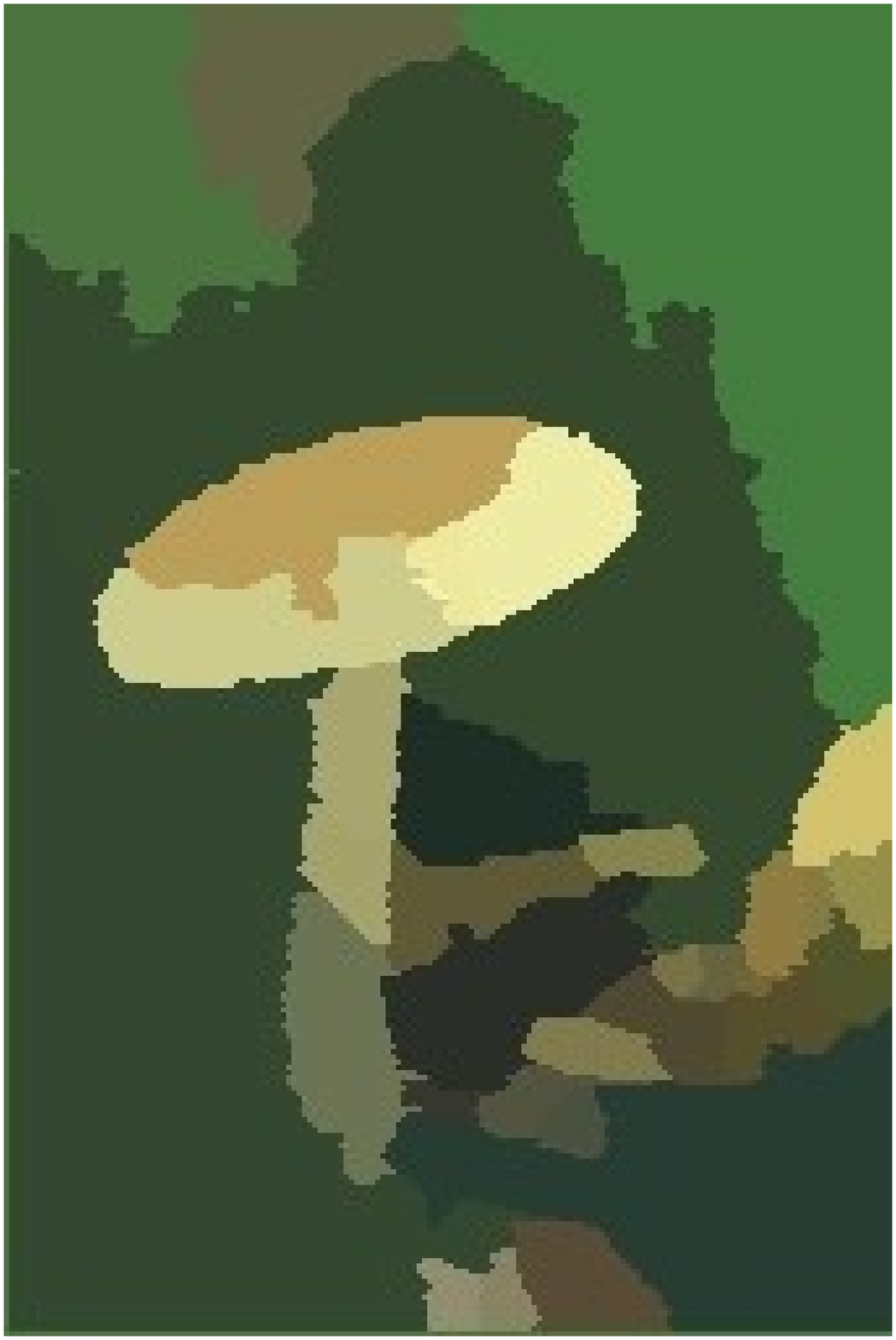} & \includegraphics[width=0.11\linewidth]{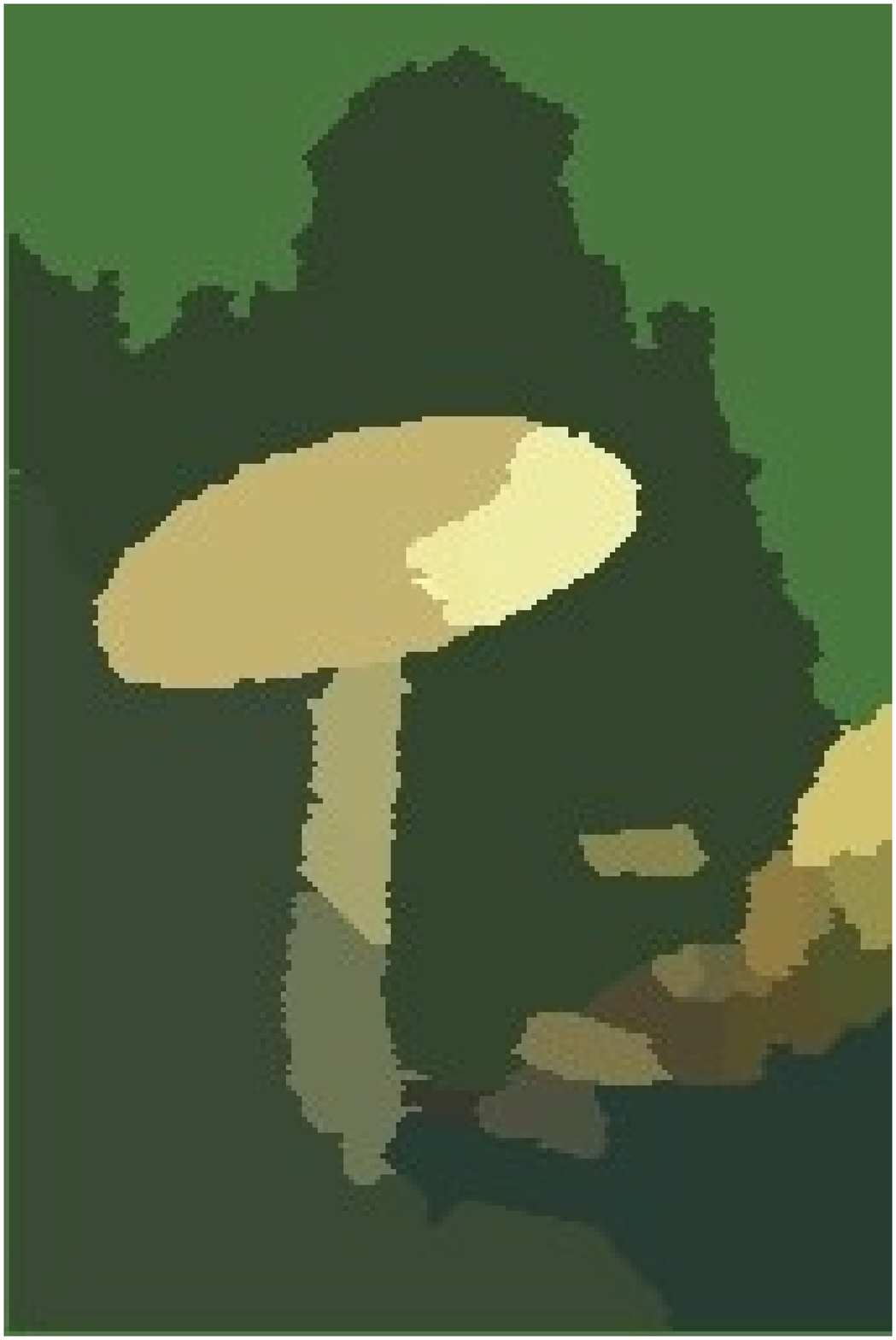}& \includegraphics[width=0.11\linewidth]{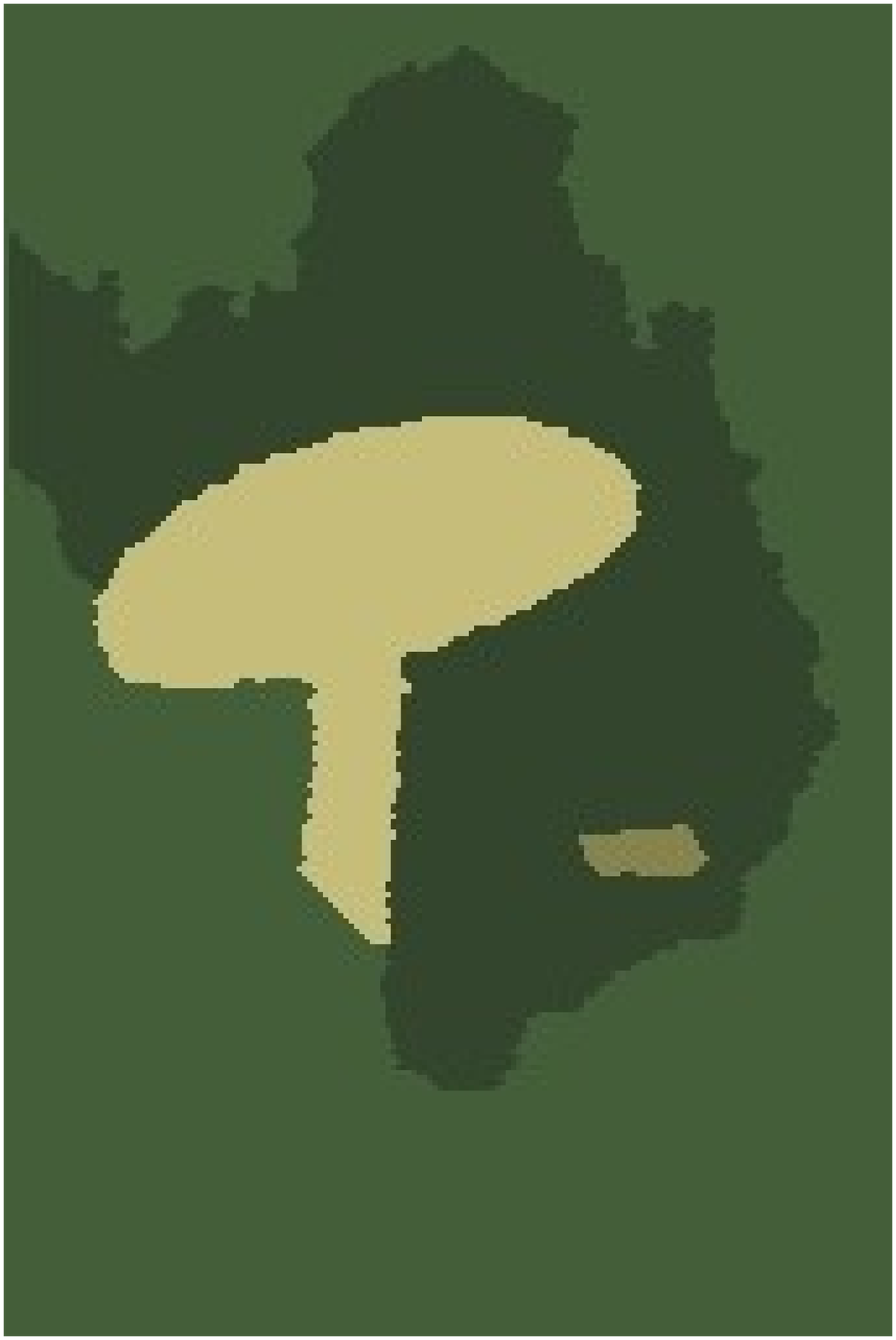}&\includegraphics[width=0.11\linewidth]{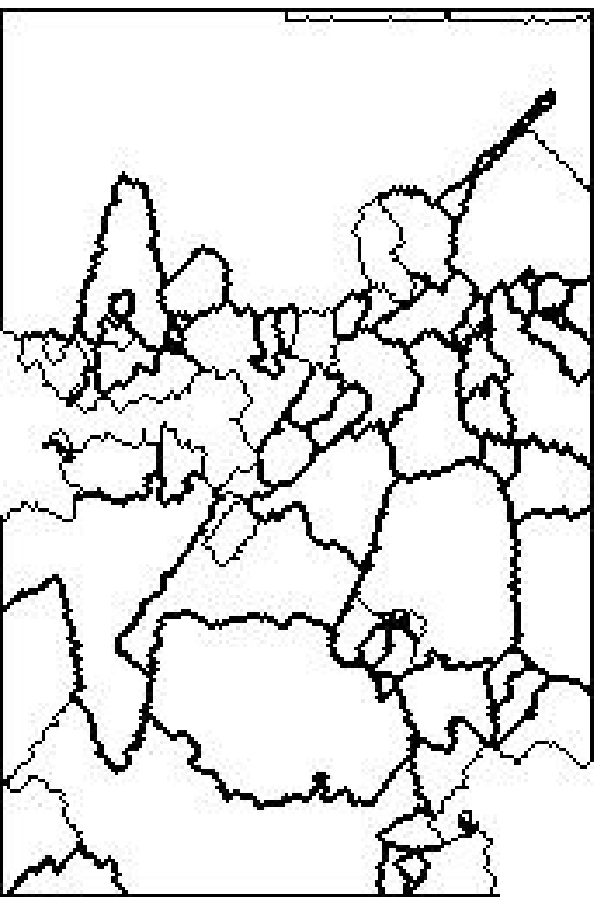} & \includegraphics[width=0.11\linewidth]{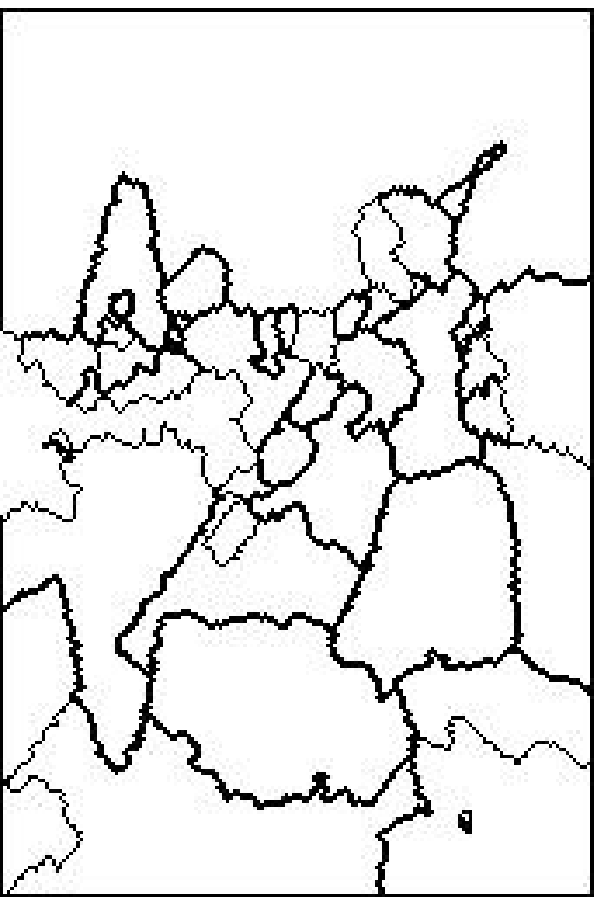} & \includegraphics[width=0.11\linewidth]{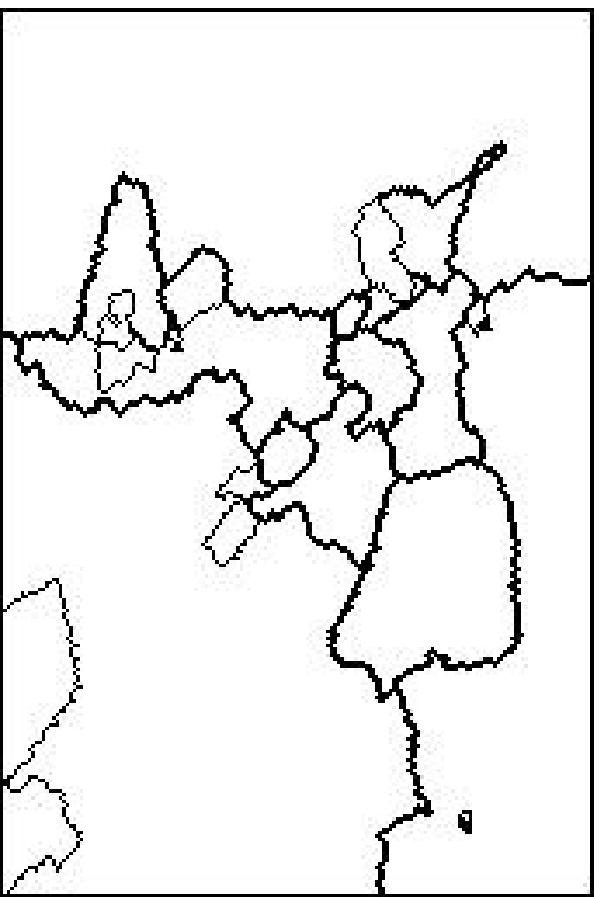}& \includegraphics[width=0.11\linewidth]{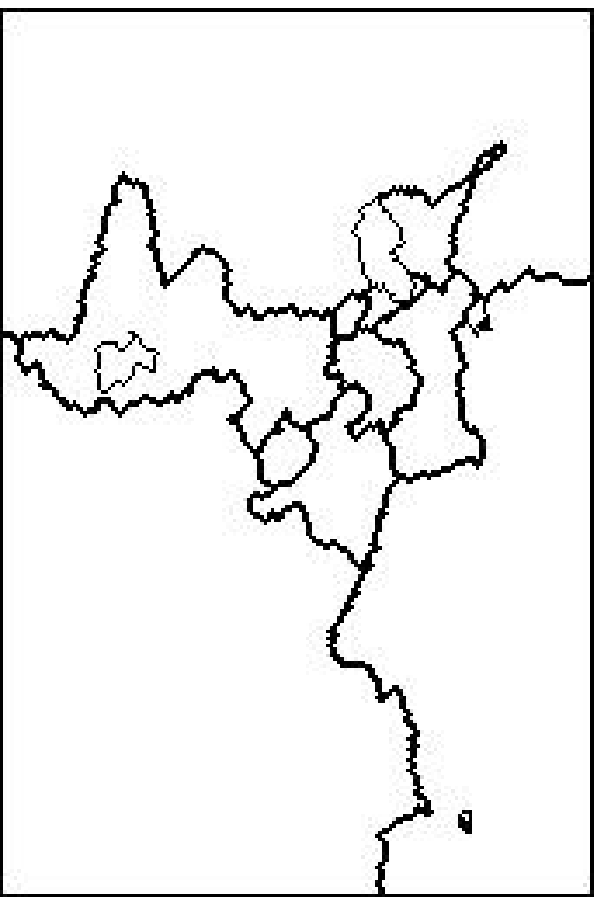}\\ 
 
\begin{minipage}[b]{0.05\linewidth} 
  $SM^2$ 
  \\ 
  \vspace{4mm} 
\end{minipage} 
&  \includegraphics[width=0.11\linewidth]{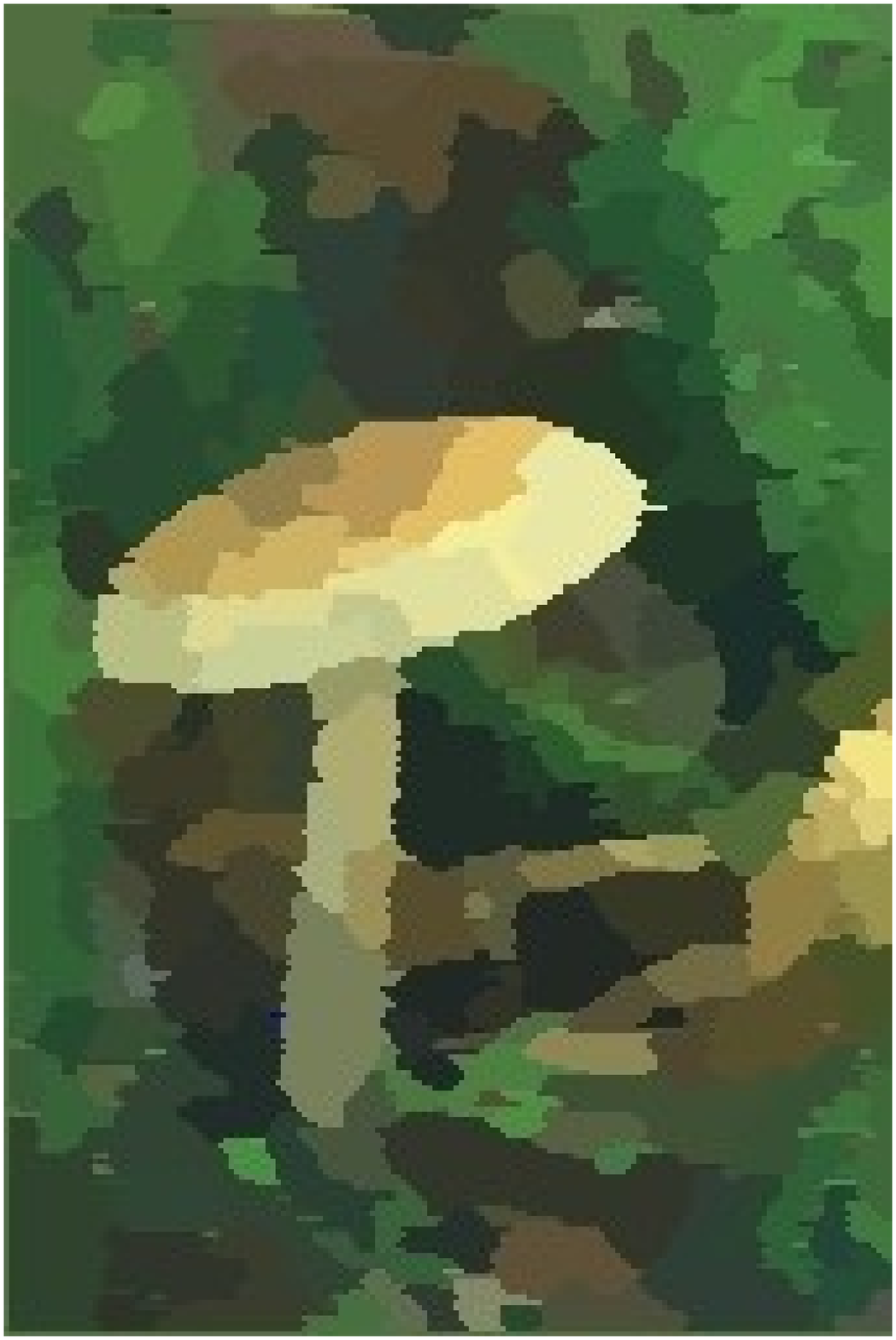} & \includegraphics[width=0.11\linewidth]{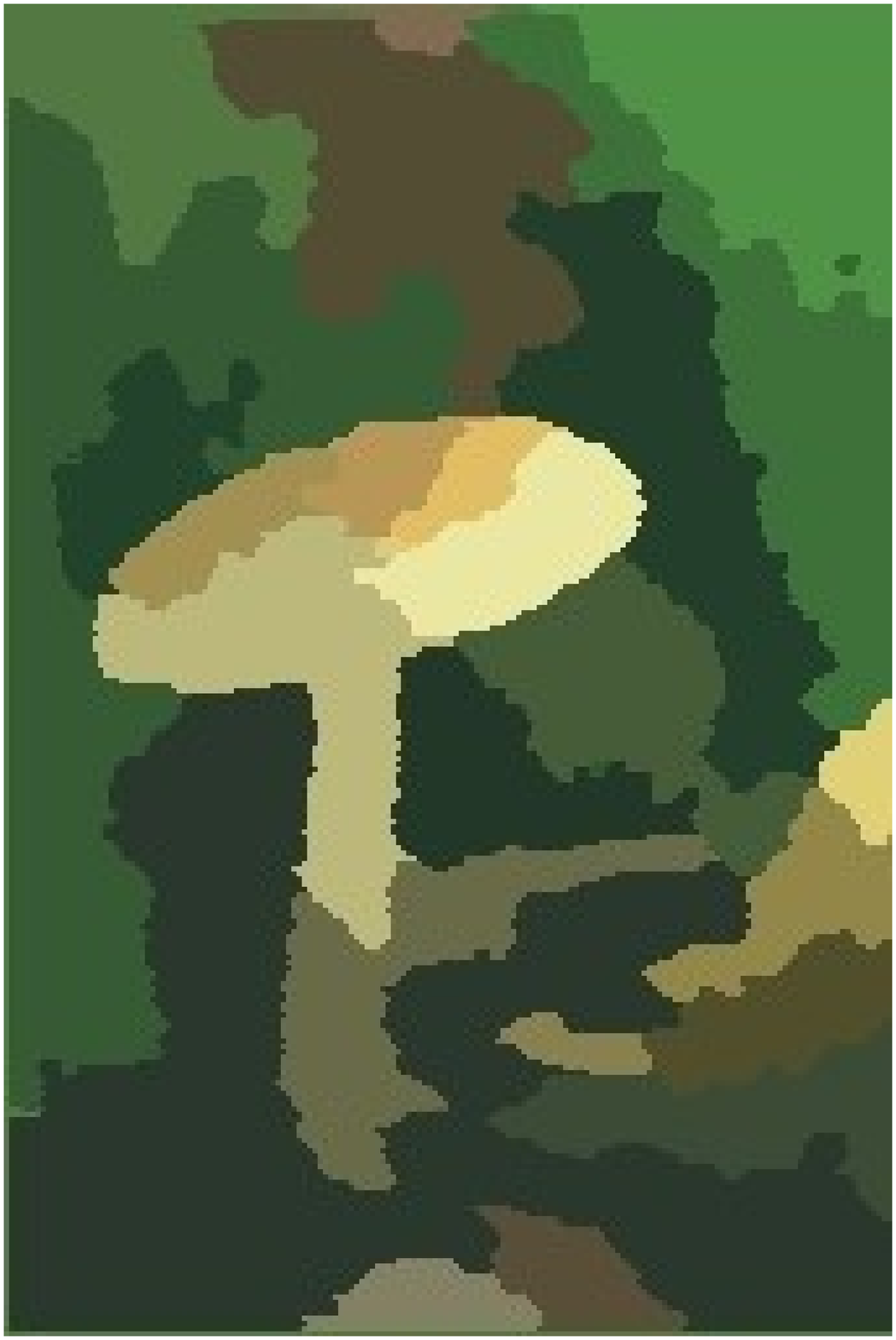} & \includegraphics[width=0.11\linewidth]{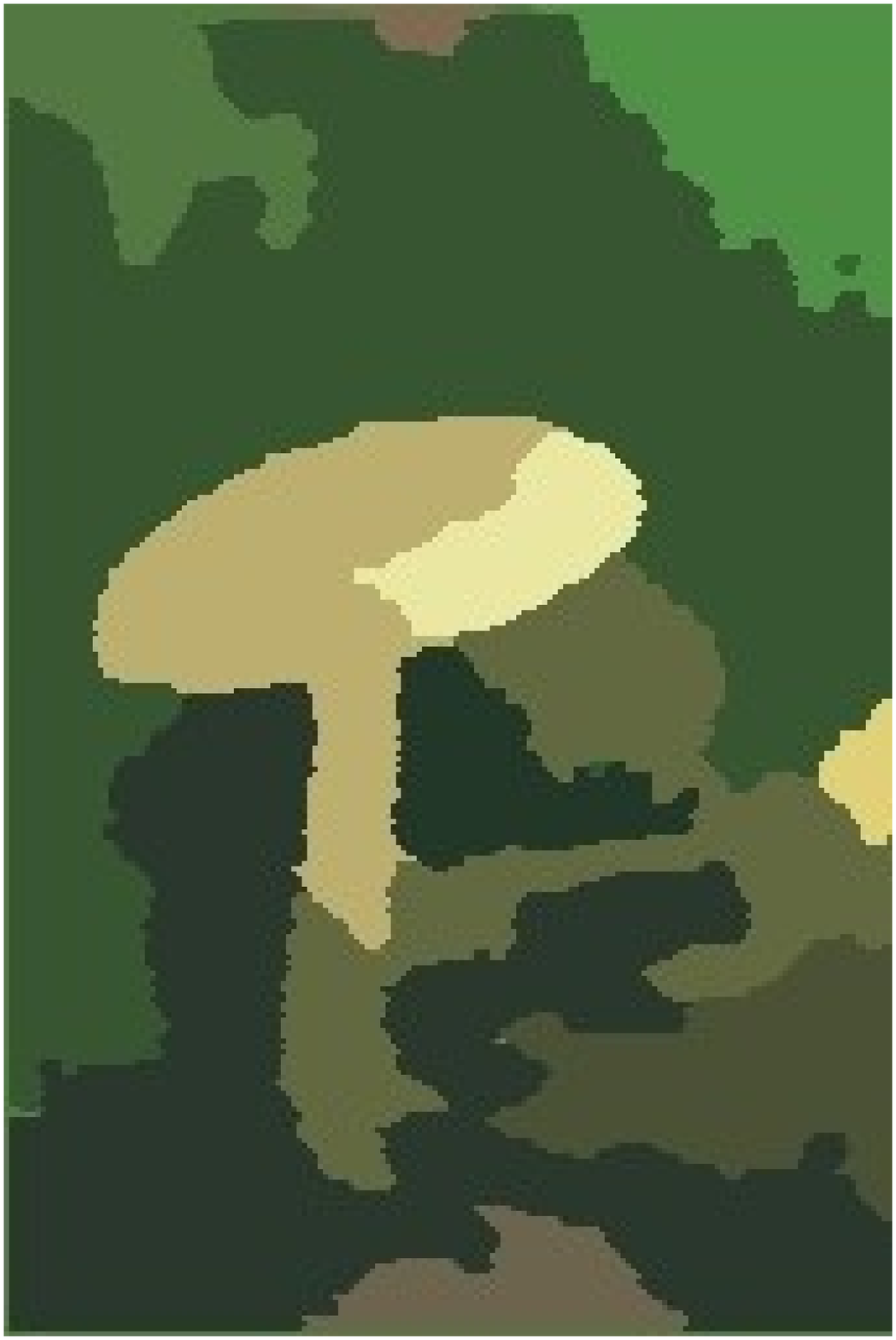}& \includegraphics[width=0.11\linewidth]{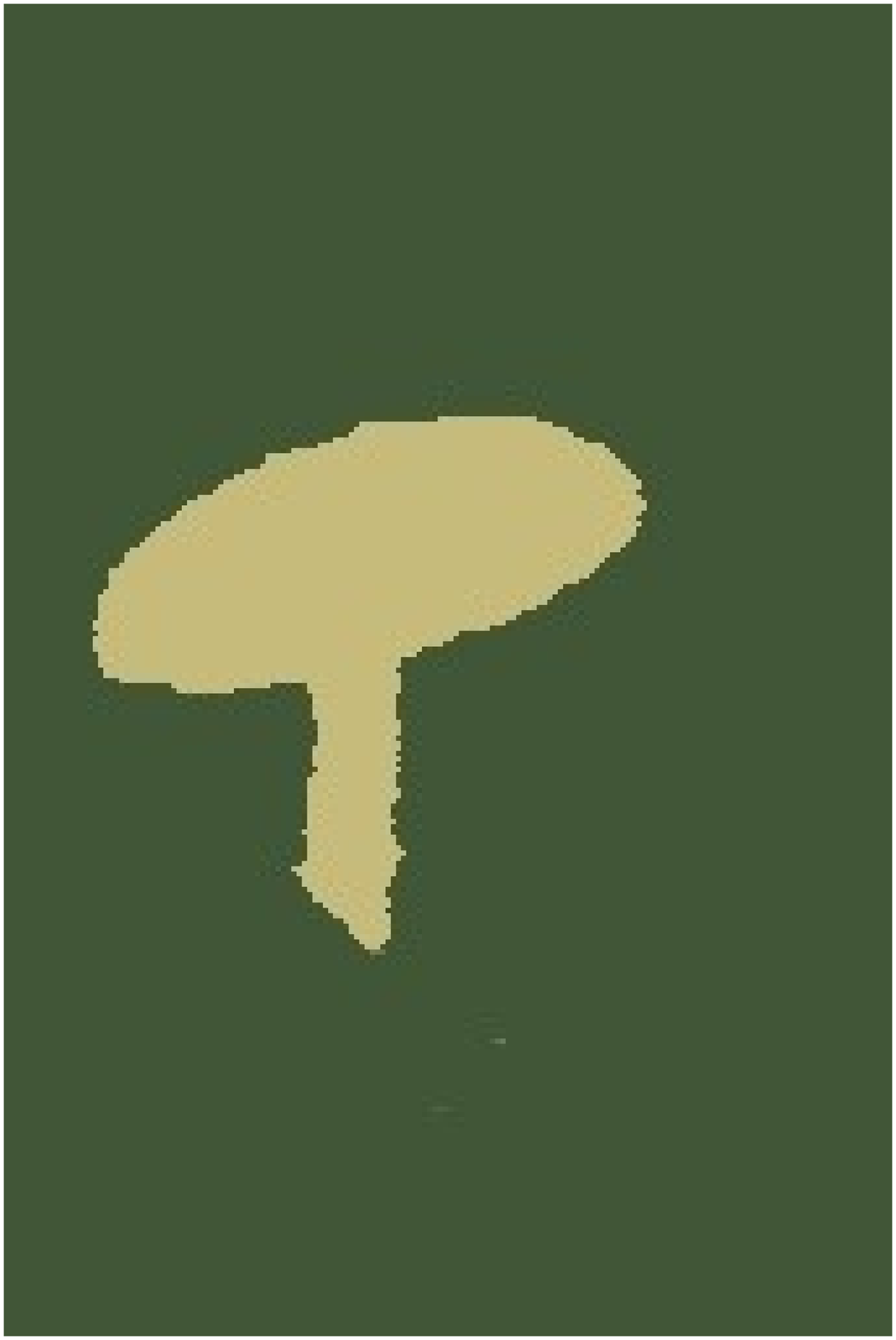}& \includegraphics[width=0.11\linewidth]{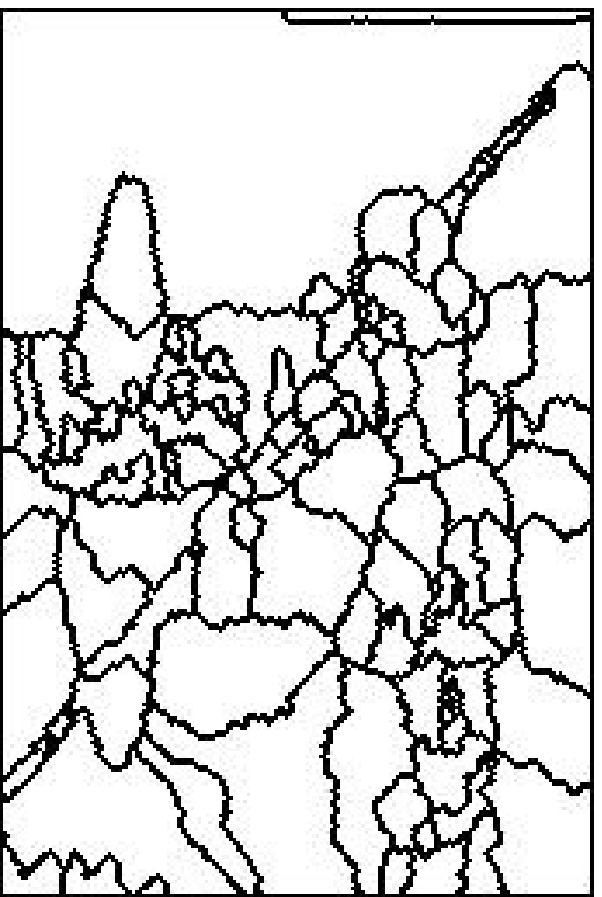} & \includegraphics[width=0.11\linewidth]{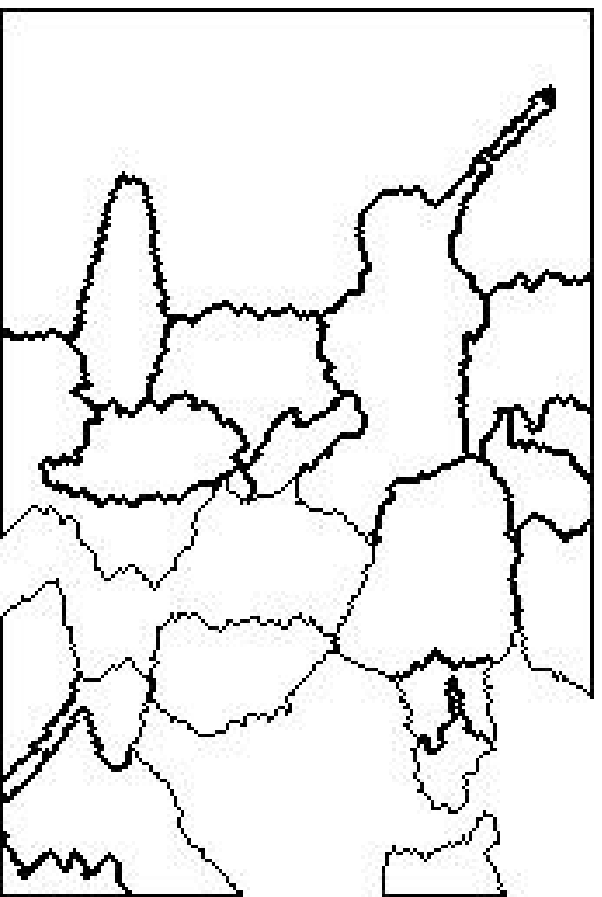} & \includegraphics[width=0.11\linewidth]{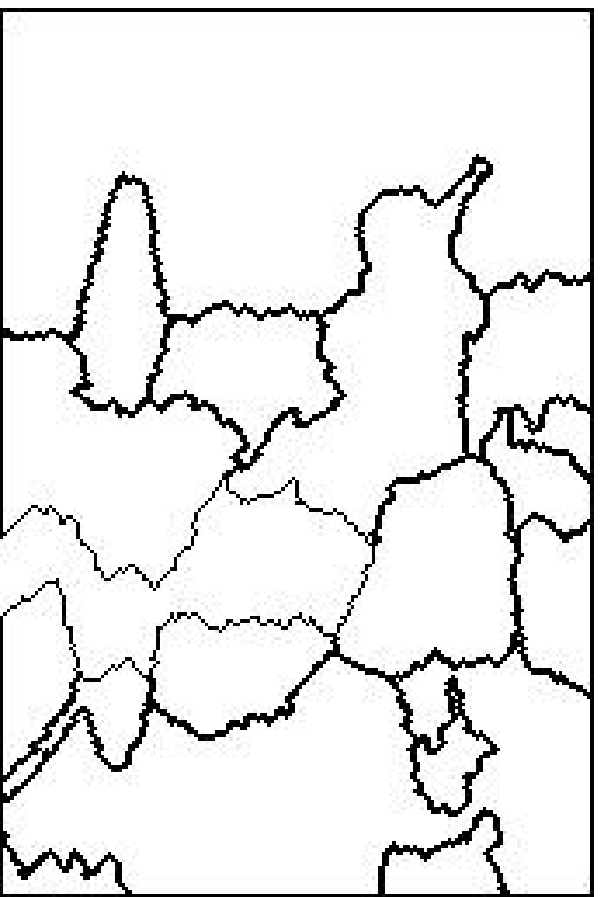}& \includegraphics[width=0.11\linewidth]{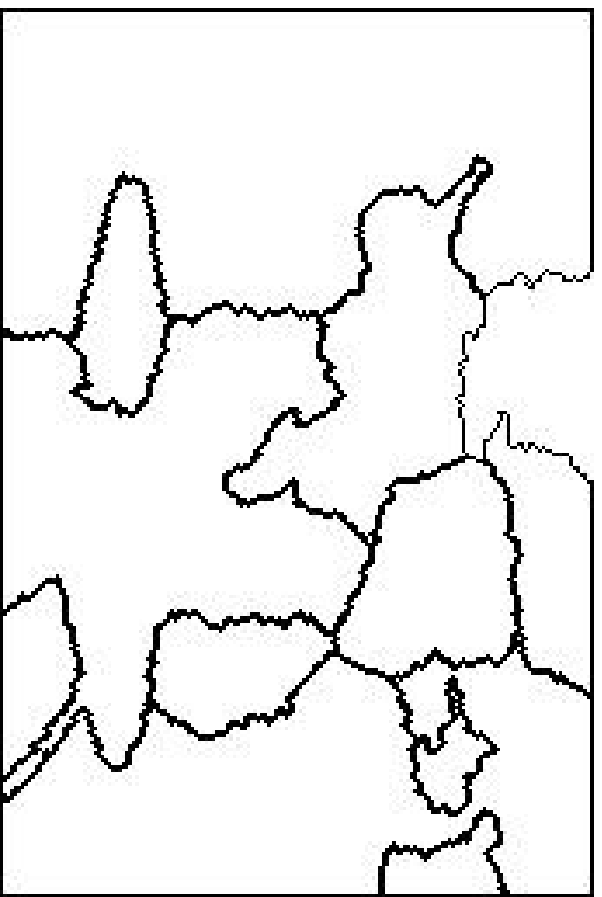}\\ 
 
\begin{minipage}[b]{0.05\linewidth} 
  SM 
  \\ 
  \vspace{4mm} 
\end{minipage} 
& 
\includegraphics[width=0.11\linewidth]{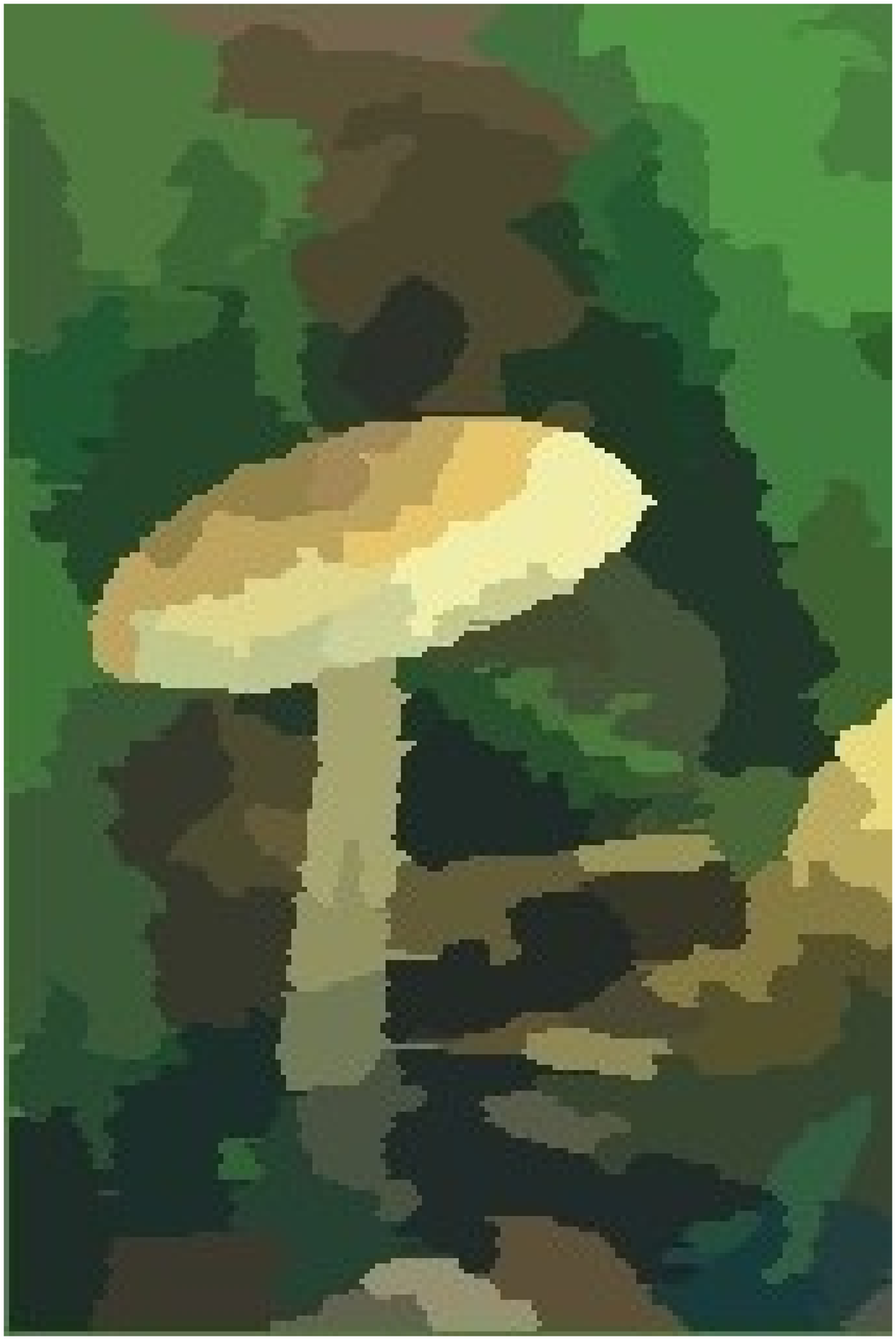}& 
\includegraphics[width=0.11\linewidth]{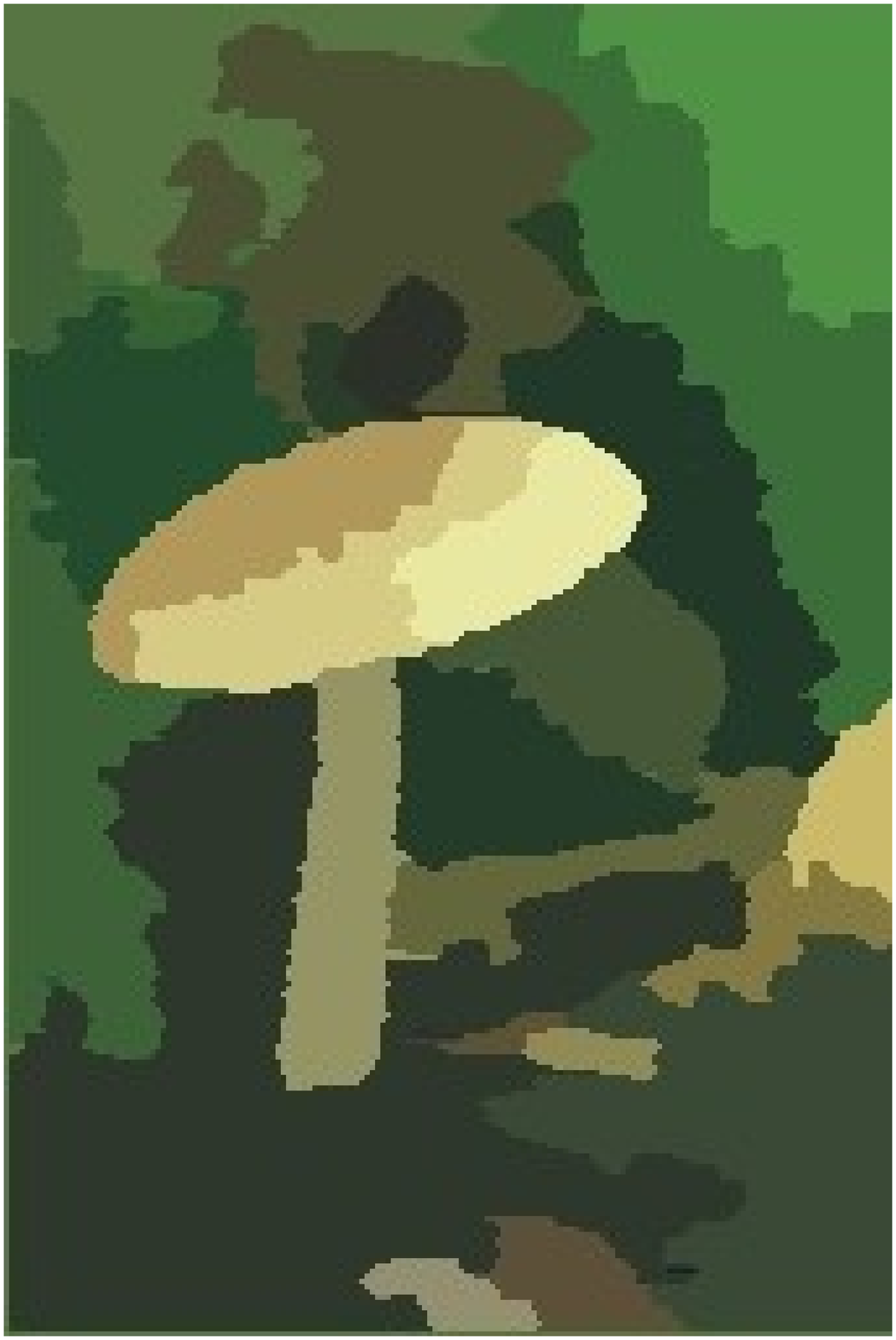} & 
\includegraphics[width=0.11\linewidth]{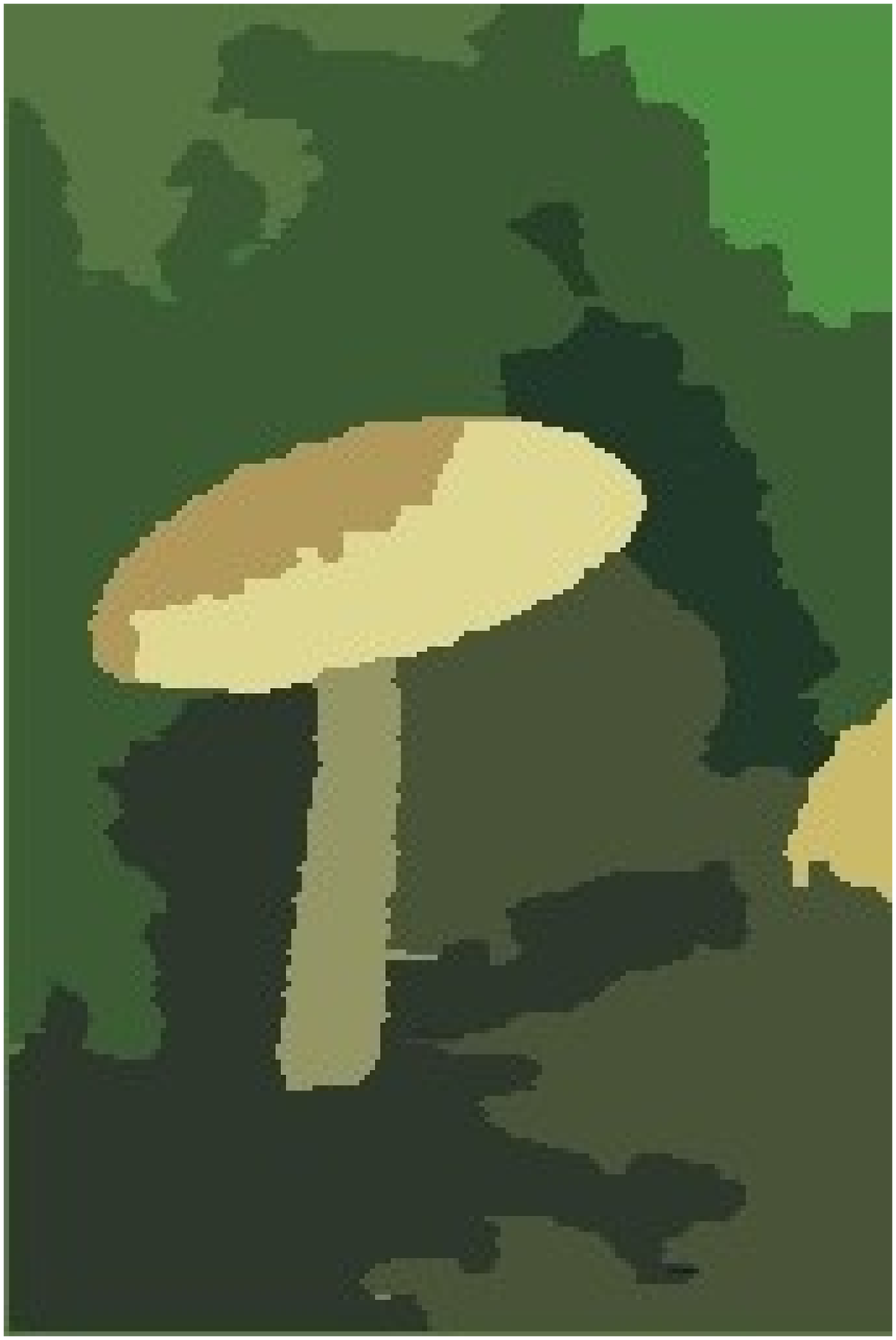}& 
\includegraphics[width=0.11\linewidth]{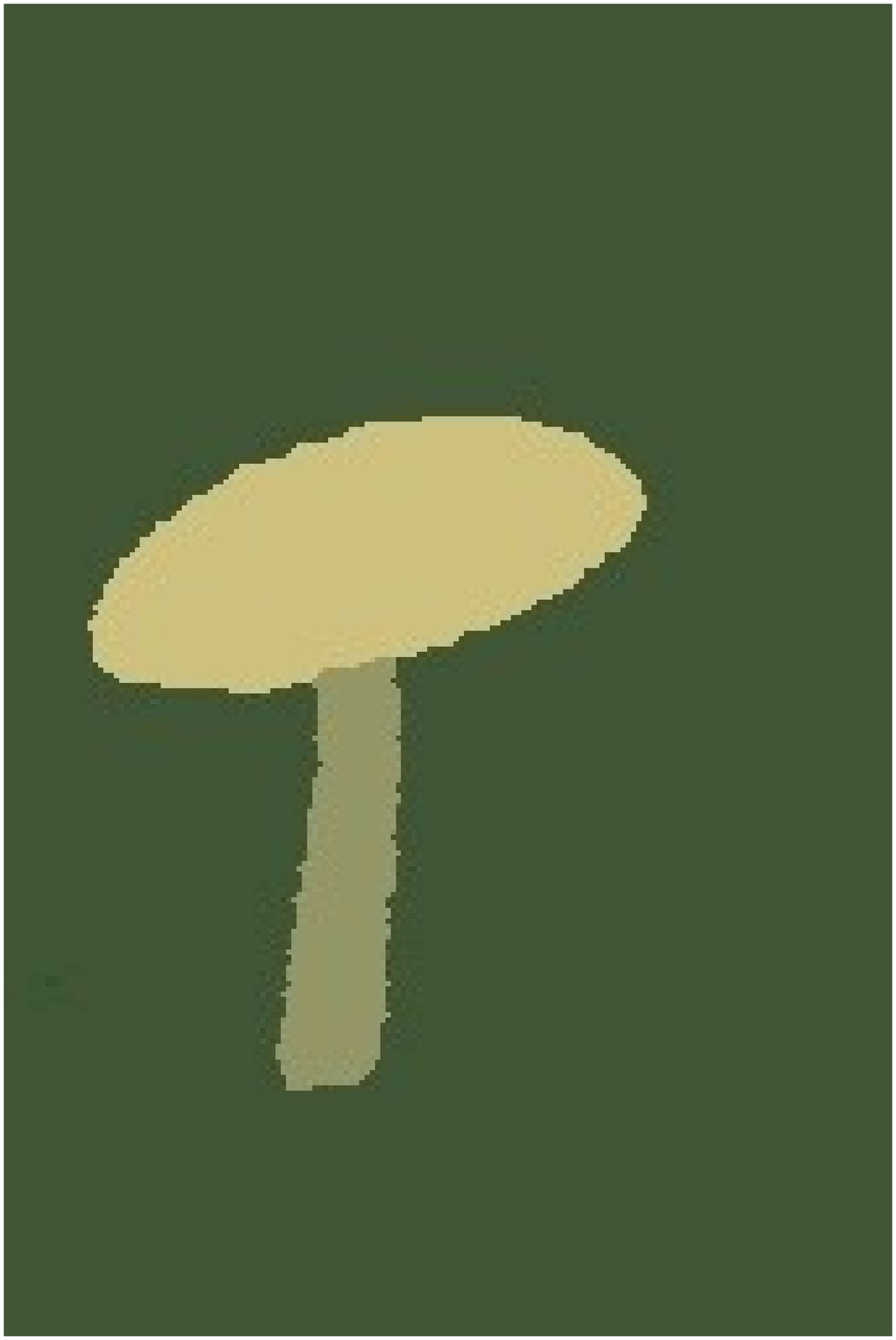}&   \includegraphics[width=0.11\linewidth]{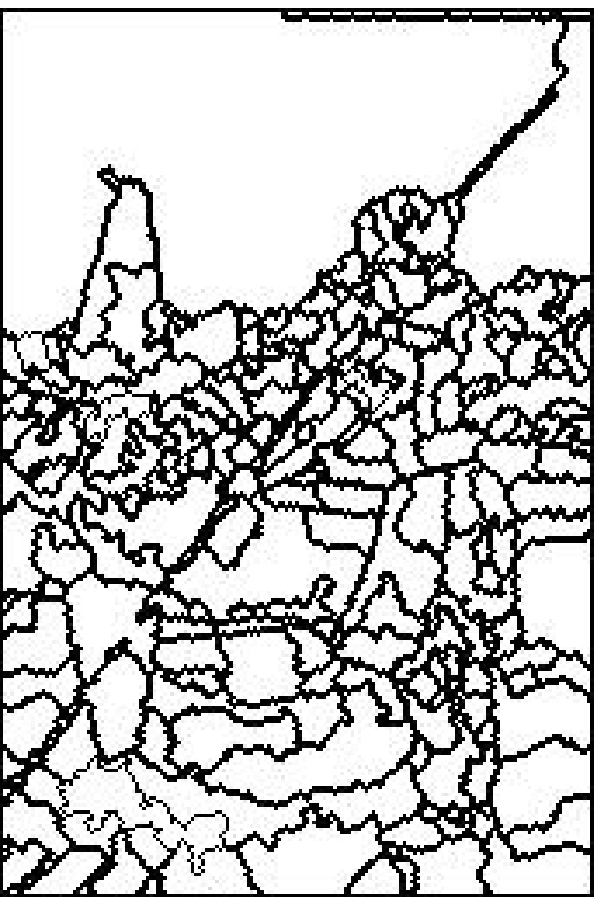} & \includegraphics[width=0.11\linewidth]{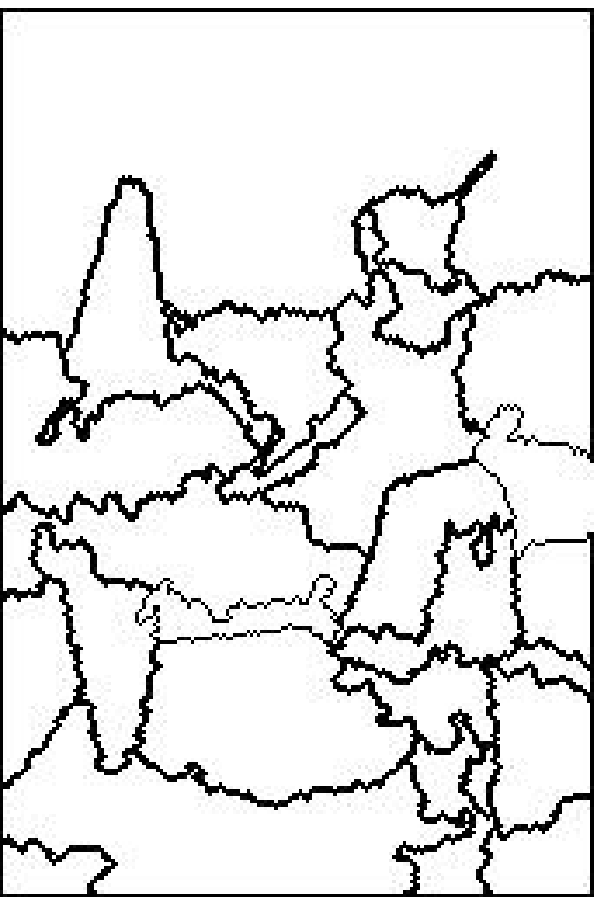} & \includegraphics[width=0.11\linewidth]{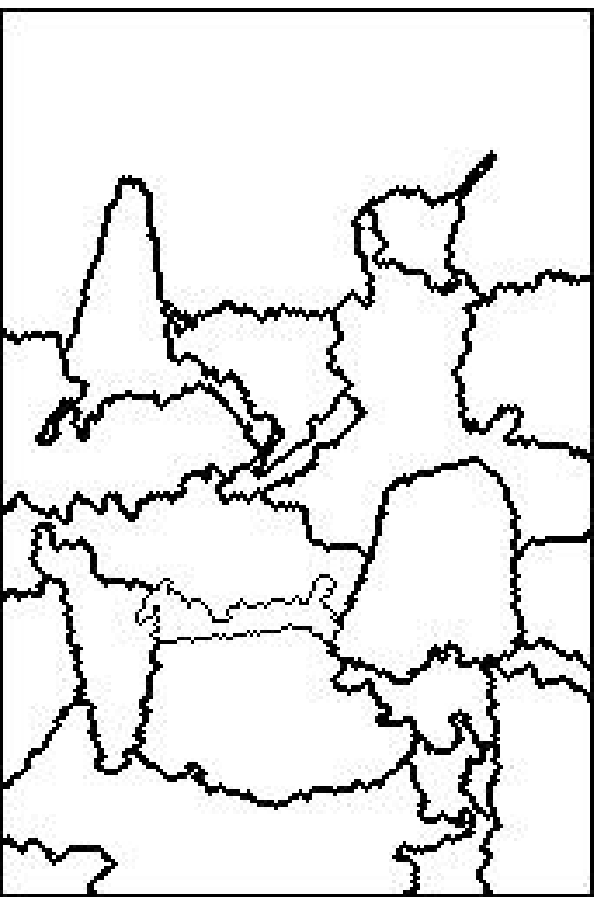}& \includegraphics[width=0.11\linewidth]{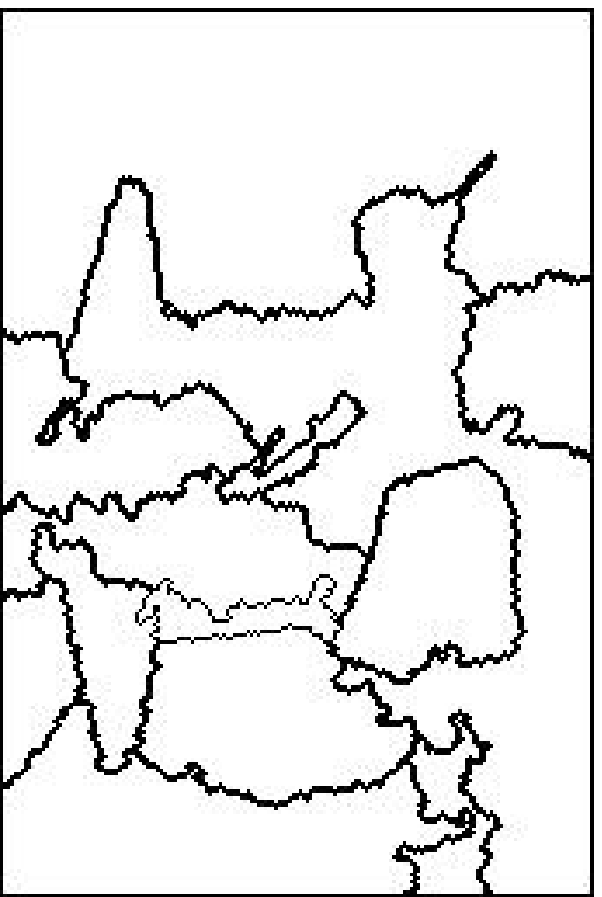}\\ 
\begin{minipage}[b]{0.05\linewidth} 
  $SM^5$ 
  \\ 
  \vspace{4mm} 
\end{minipage}
&   \includegraphics[width=0.11\linewidth]{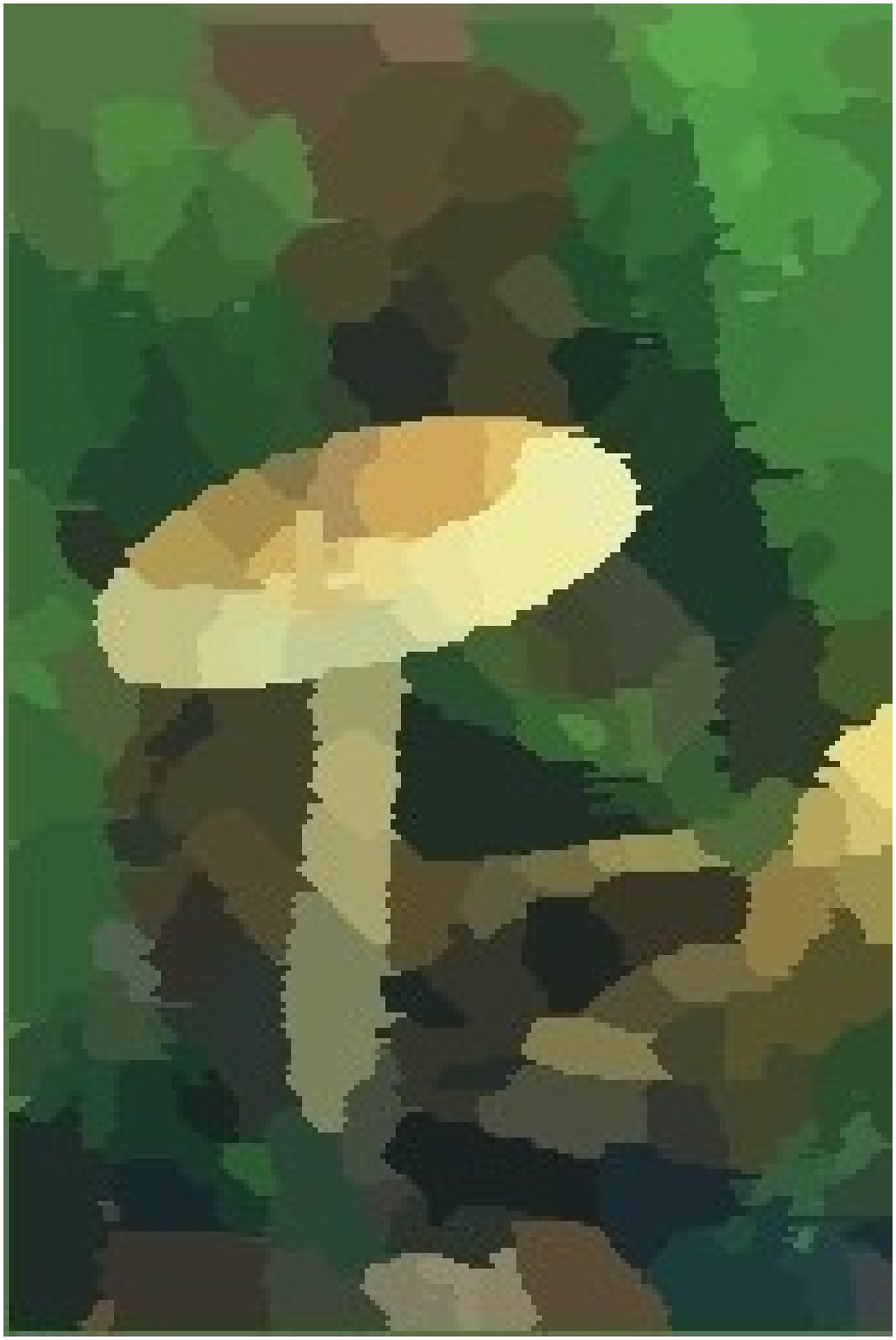} & \includegraphics[width=0.11\linewidth]{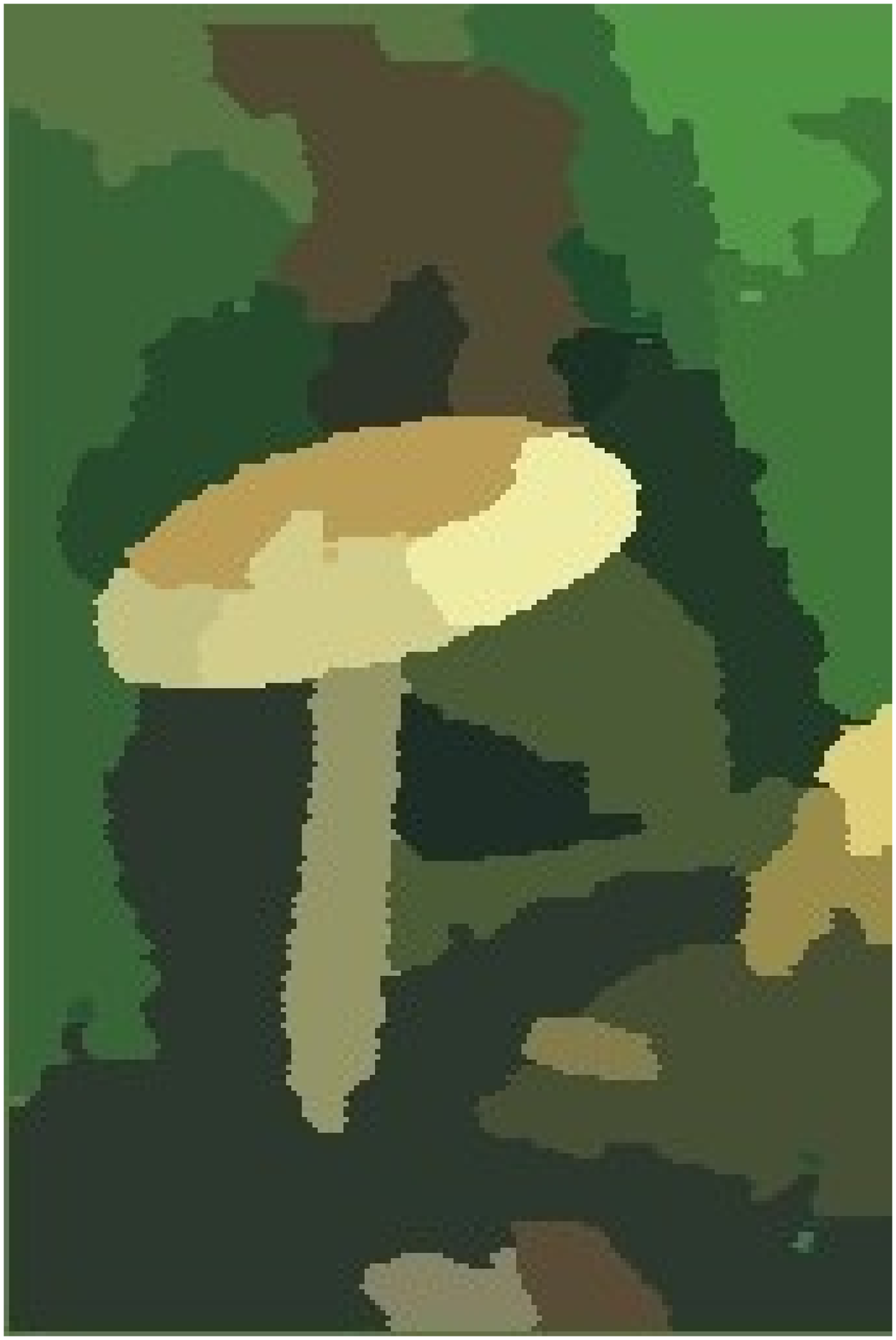} & \includegraphics[width=0.11\linewidth]{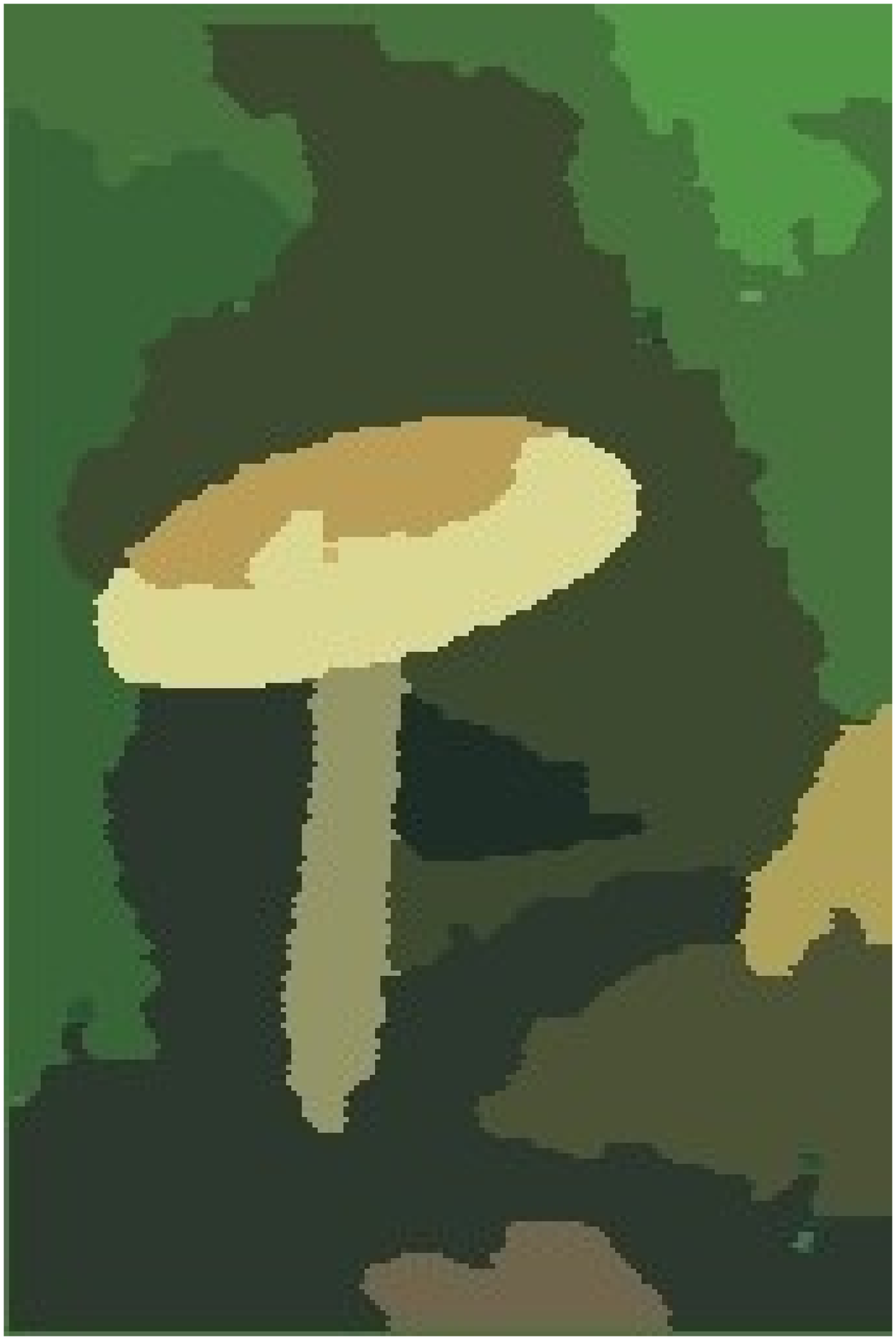}& \includegraphics[width=0.11\linewidth]{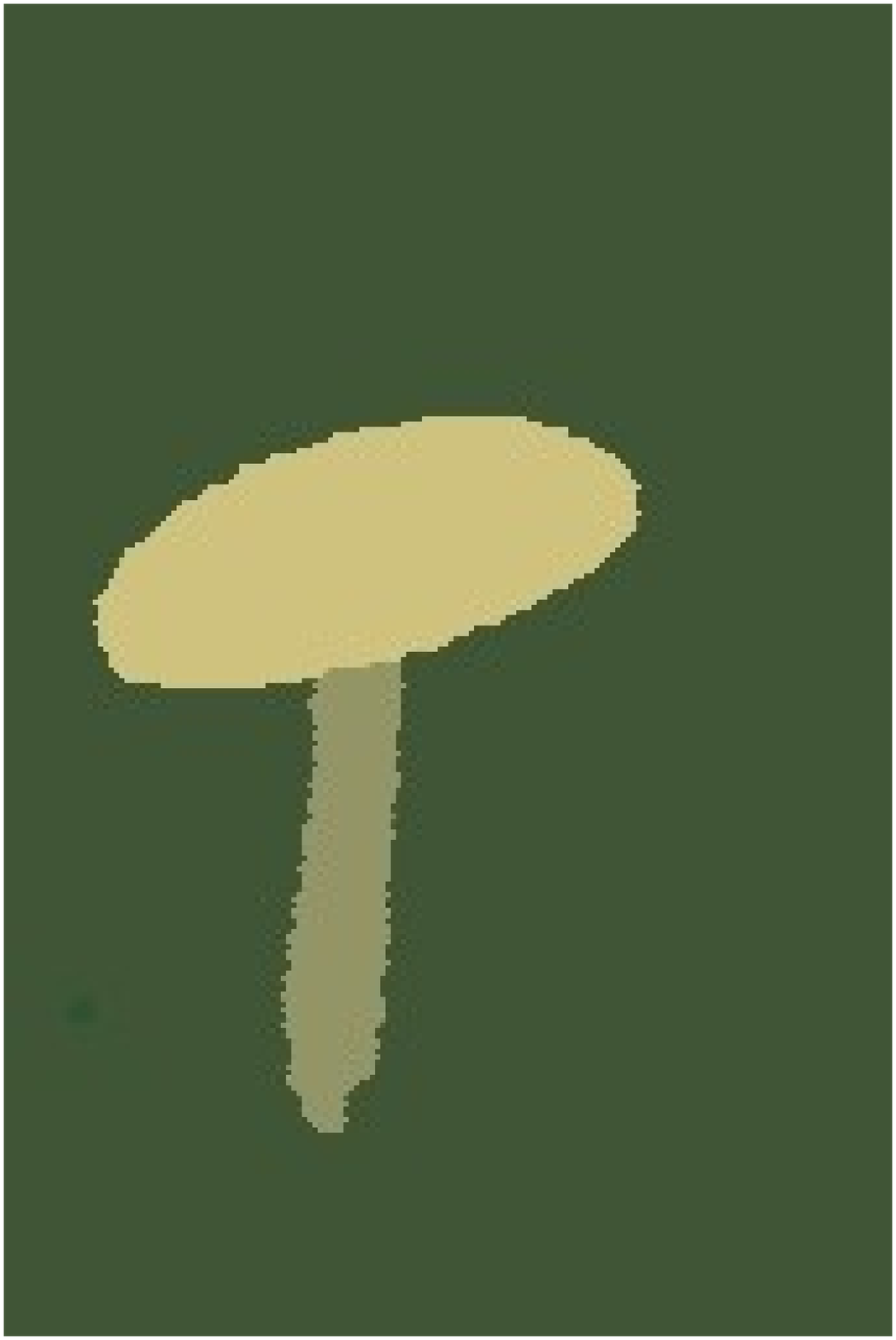}&  \includegraphics[width=0.11\linewidth]{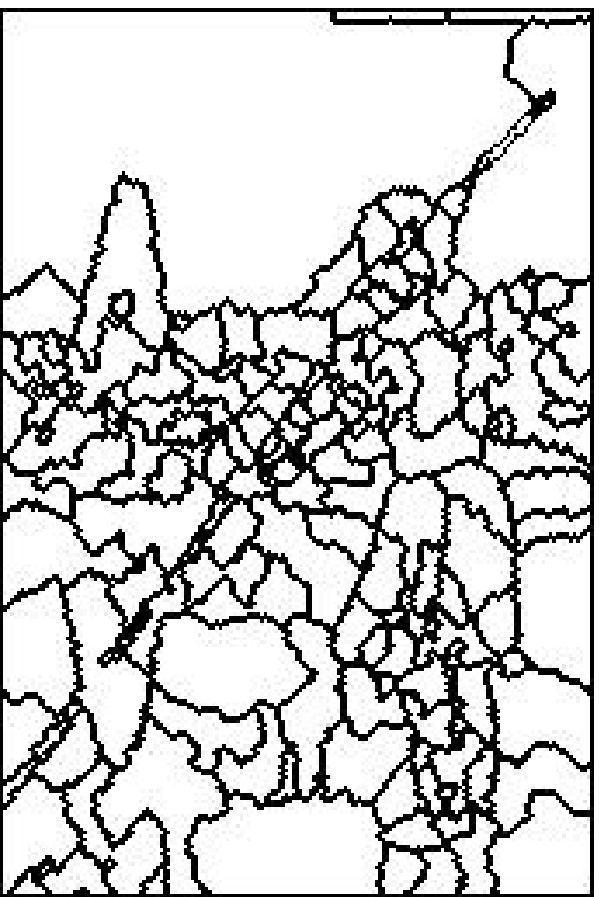} & \includegraphics[width=0.11\linewidth]{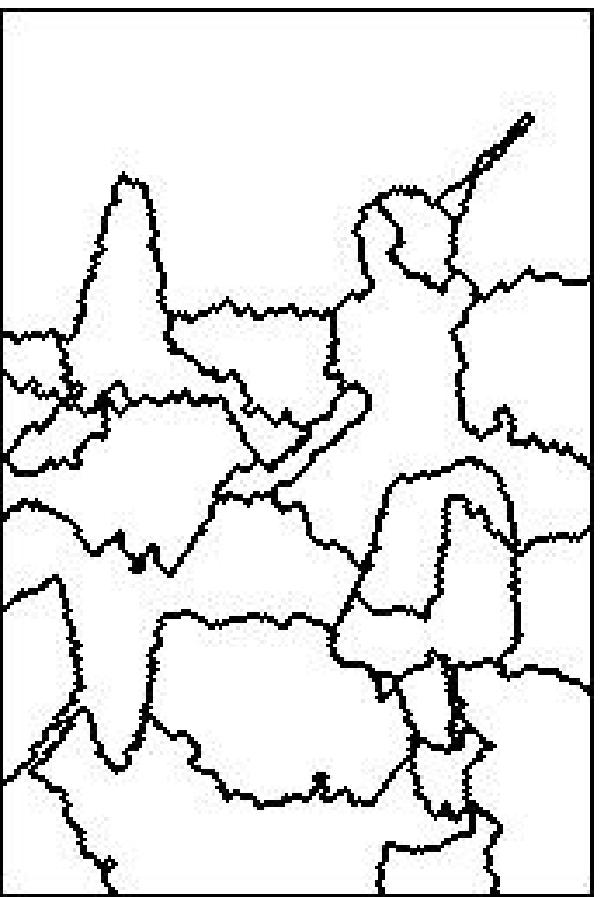} & \includegraphics[width=0.11\linewidth]{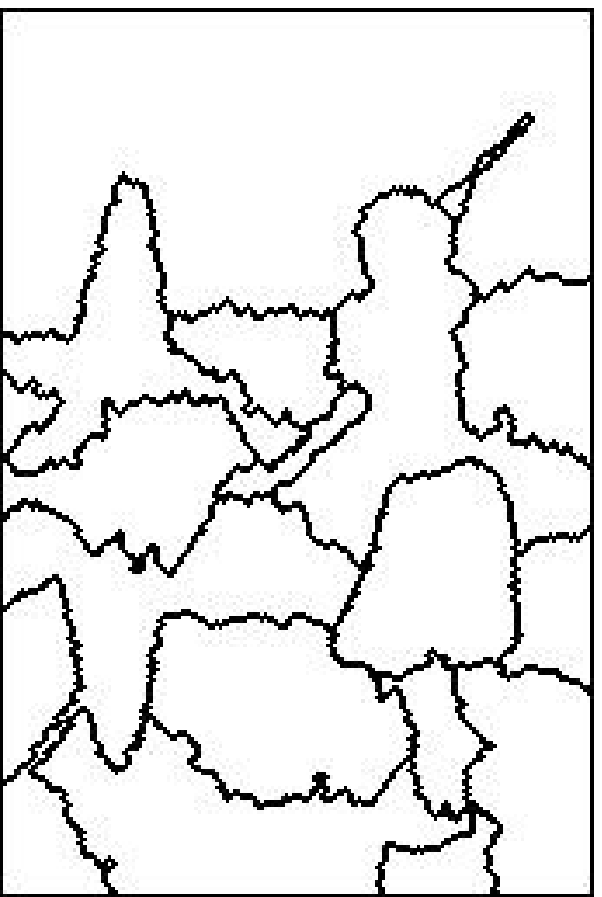}& \includegraphics[width=0.11\linewidth]{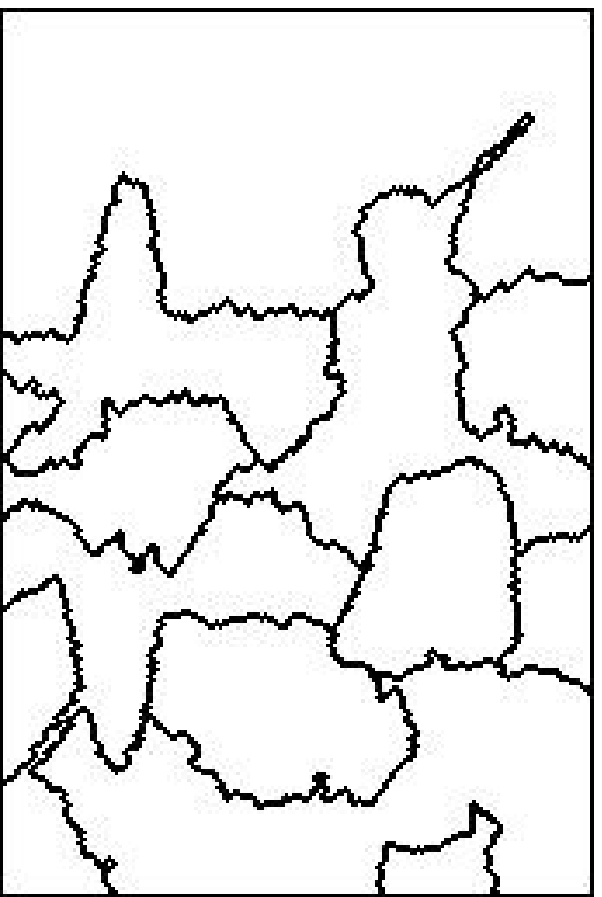}\\

~&$\lambda = 0.2$&$\lambda = 0.4$&$\lambda = 0.6$&$\lambda = 0.8$&$\lambda = 0.2$&$\lambda = 0.4$&$\lambda = 0.6$&$\lambda = 0.8$ 
\end{tabular}\\ 
\caption{\label{fig:2a2} Partitions of the mushroom and the fisherman images at different 
  scales. Each line of the array corresponds to an heuristic whose 
  acronym is indicated on the first column.} 
\end{figure}

\begin{figure} 
 
\mbox{ }\hfill 
\subfigure [Execution time]{\includegraphics[width=3.5cm]{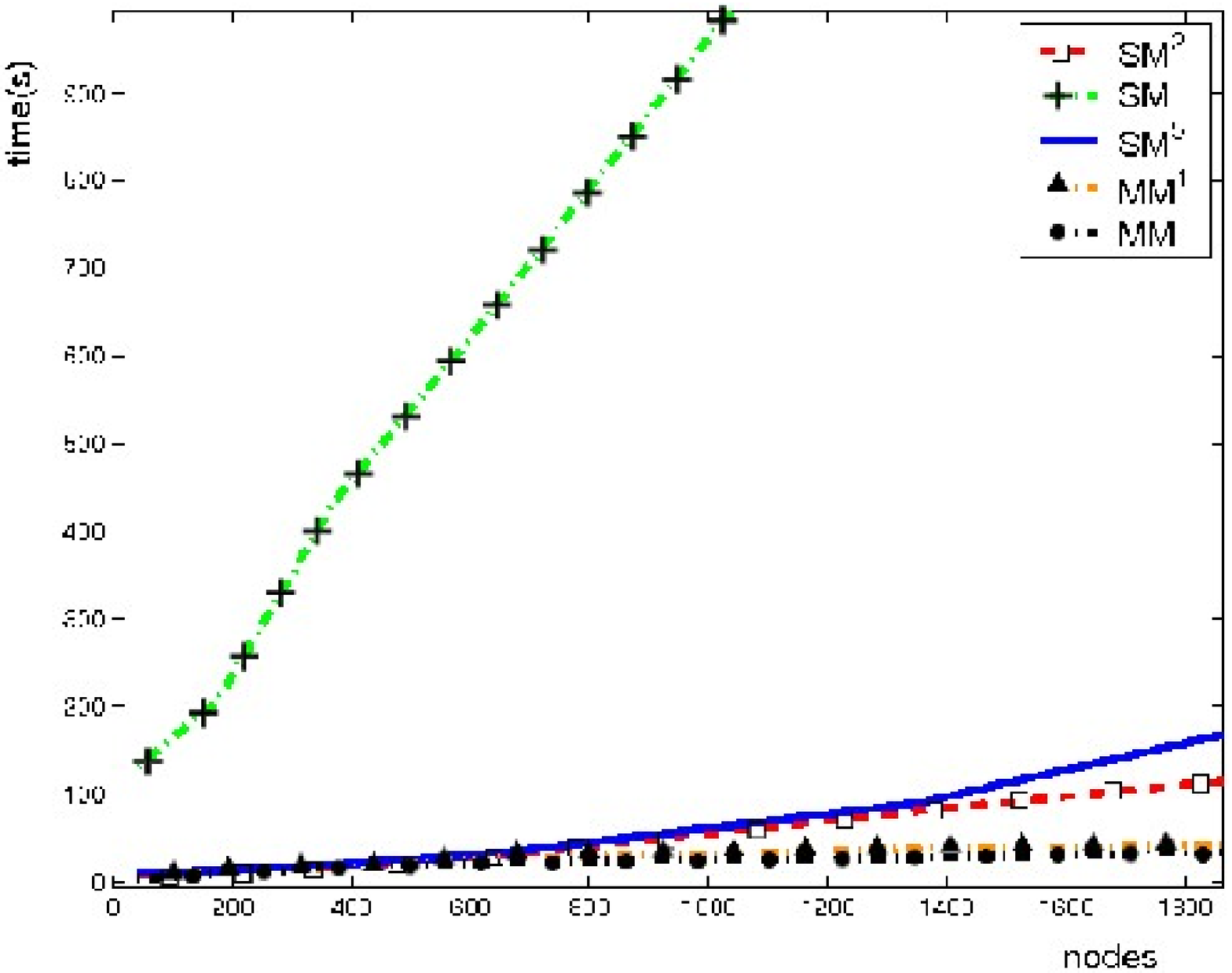}} 
\hfill 
 \subfigure[$E_\lambda(C^*_\lambda(H))$] {\includegraphics[width=3.5cm]{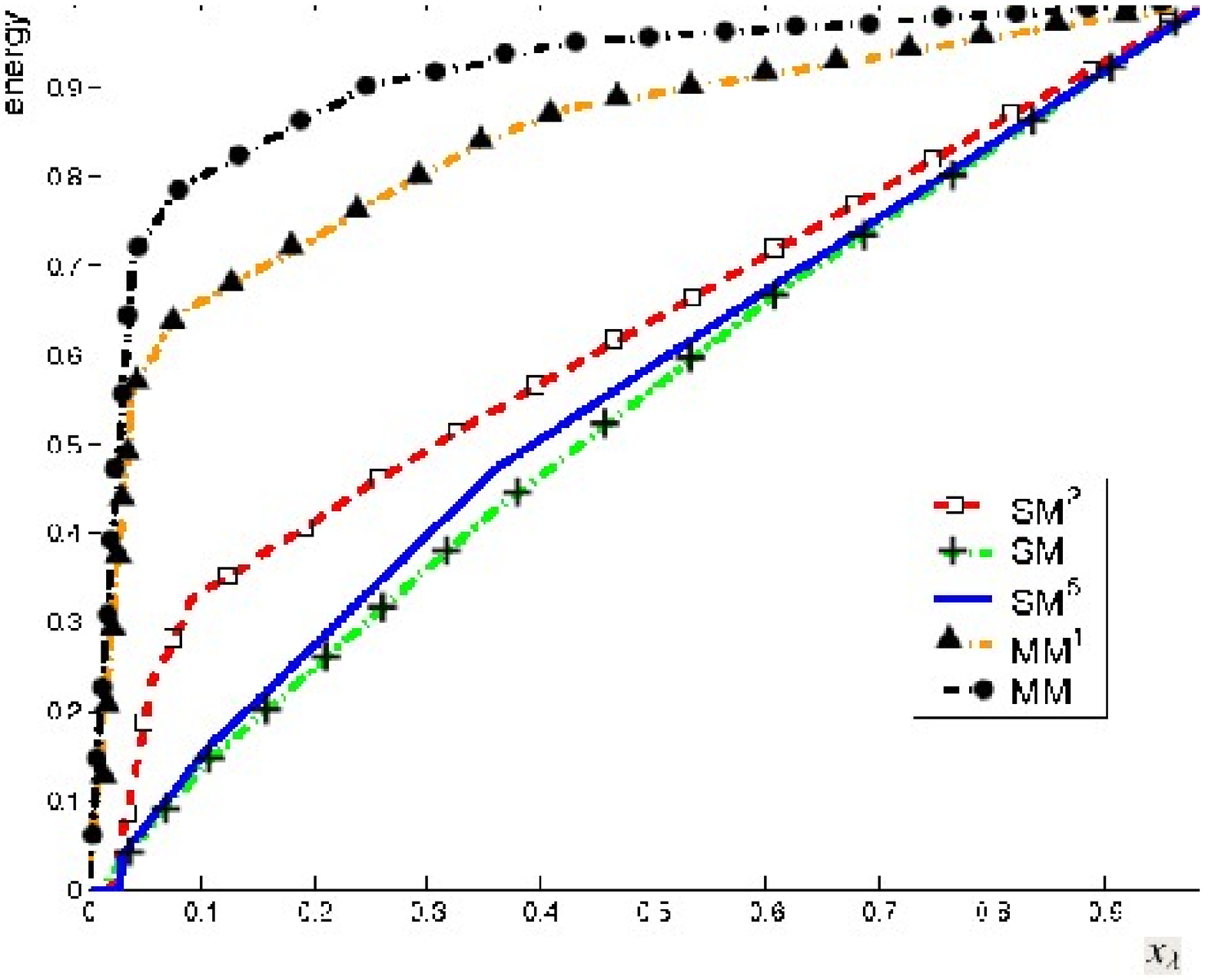}} 
\hfill 
\subfigure[Energy's Bounds]
{ 
 \begin{minipage}{3cm} 
  \unitlength .6mm 
    \begin{picture}(60,8) 
      \tiny  
      \put(10,10){\vector(1,0){40}} 
      \put(10,10){\vector(0,1){40}} 
      \drawline(10,20)(20,30)(30,35)(40,37)  
      \dashline{3}(10,31)(40,37) 
      \dashline{3}(10,10)(40,37)  
      \dashline{4}(40,10)(40,37) 
      \put(38,5){$\lambda_{max}$} \put(43,23){$E_{\lambda_{max}}(P_{max})$} 
      \put(15,35){$E_{\lambda}(P_{max})$} \put(0,20){$E_0(P_0)$} 
      \put(15,45){$E_\lambda(C^*(H))$} \put(55,5){$\lambda$} 
    \end{picture} 
 \end{minipage} 
} 
\hfill   \mbox{\nobreakspace}   
\caption{(a) execution times of the different heuristics on the 
  Mushroom image (Fig.~\protect\ref{fig:2a2}) using an initial 
  partition with a varying number of regions. (b) mean energies of 
  optimal cuts obtained by our heuristics on the Berkeley database. 
  (c) bounds of the optimal cut's energies.} 
  \label{fig:curves} 
\end{figure} 
 
The different heuristics presented in this paper have been evaluated 
on the Berkeley database. The evaluated heuristics include our 
parallel merge heuristic based on a maximal matching (MM) and the 
variation of this method($MM^1$) which merges at each step the edges 
selected during the first iteration (Section~\ref{subsec:parallel}). 
We also evaluated our sequential method (SM) and two variations of 
this method: the first variation $(SM^2)$, considers for each region 
$R$ of the partition the subsets of cardinal $2$ of $V(R)$. This 
method corresponds to the heuristic proposed by Guigues. We also 
evaluated an intermediate method ($SM^5$) which restricts the cardinal 
of the subsets of $V(R)$ including $R$ to an upper threshold fixed to 
five in these experiments. All the experiments have used an initial 
partition obtained by a Watershed algorithm~\cite{brun-05}. 
 
\begin{figure}[t] 
\centering 
\mbox{ }\hfill
\subfigure[Original Images]{
\includegraphics[width=0.11\linewidth]{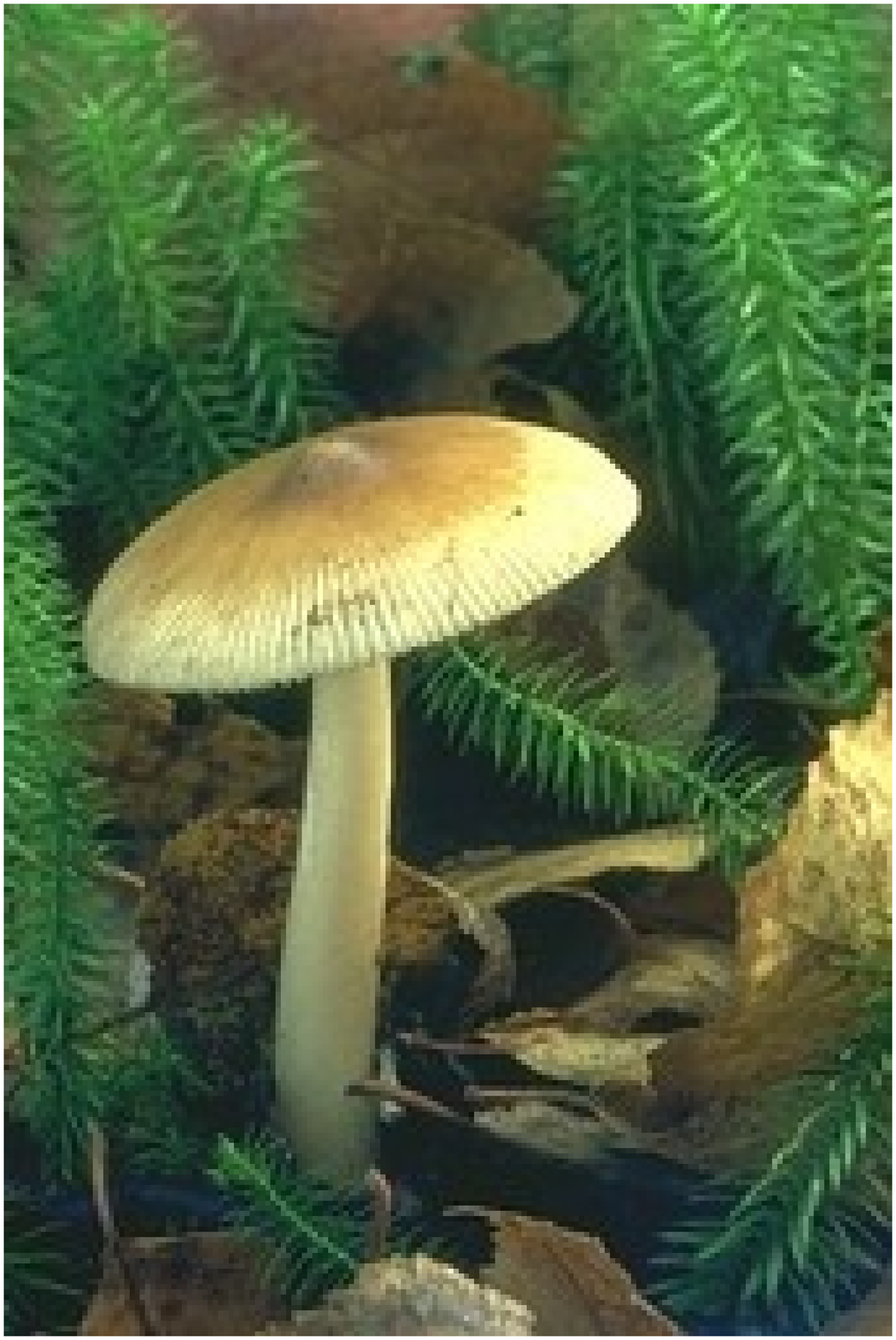}
\includegraphics[width=0.11\linewidth]{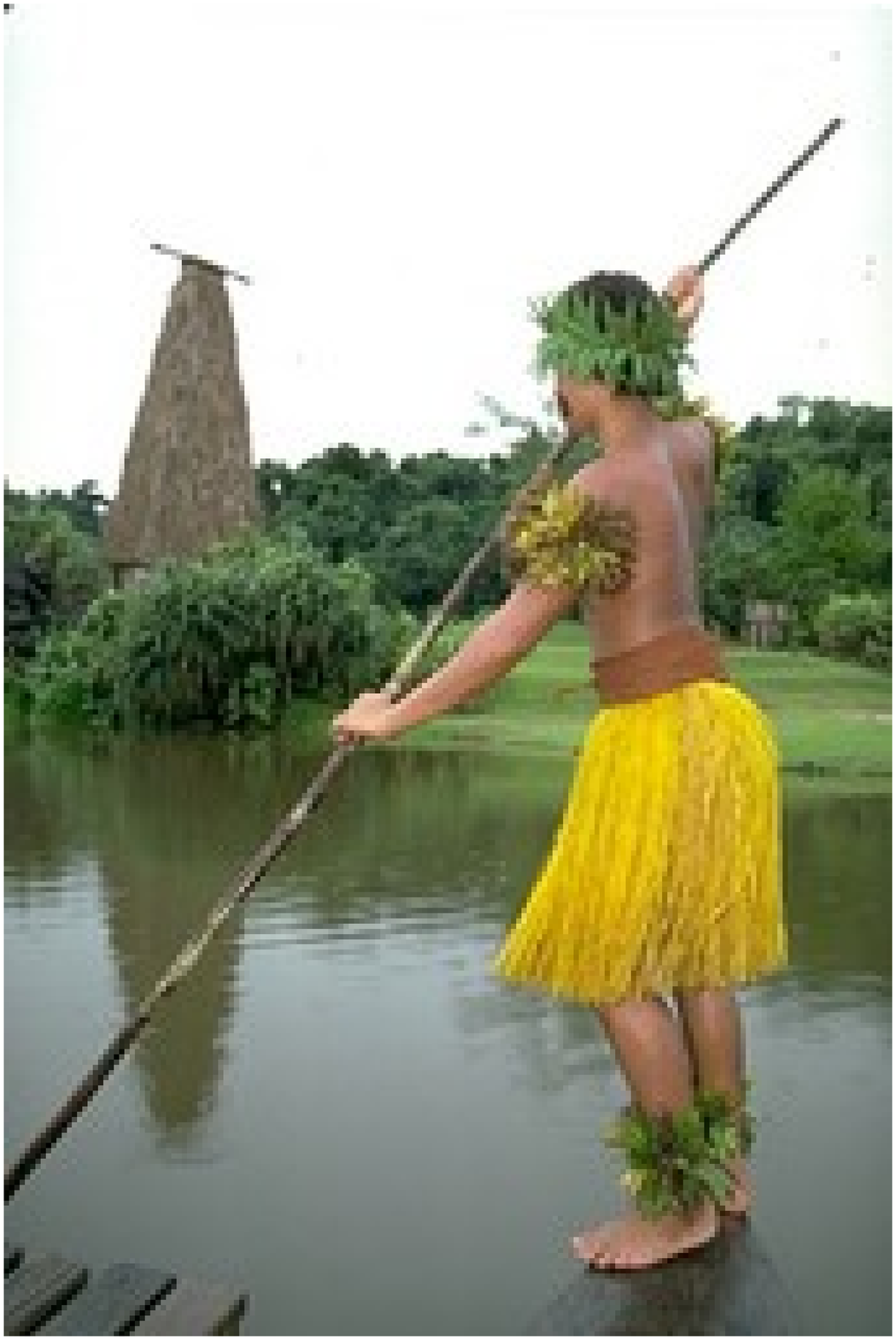}
\includegraphics[width=0.11\linewidth]{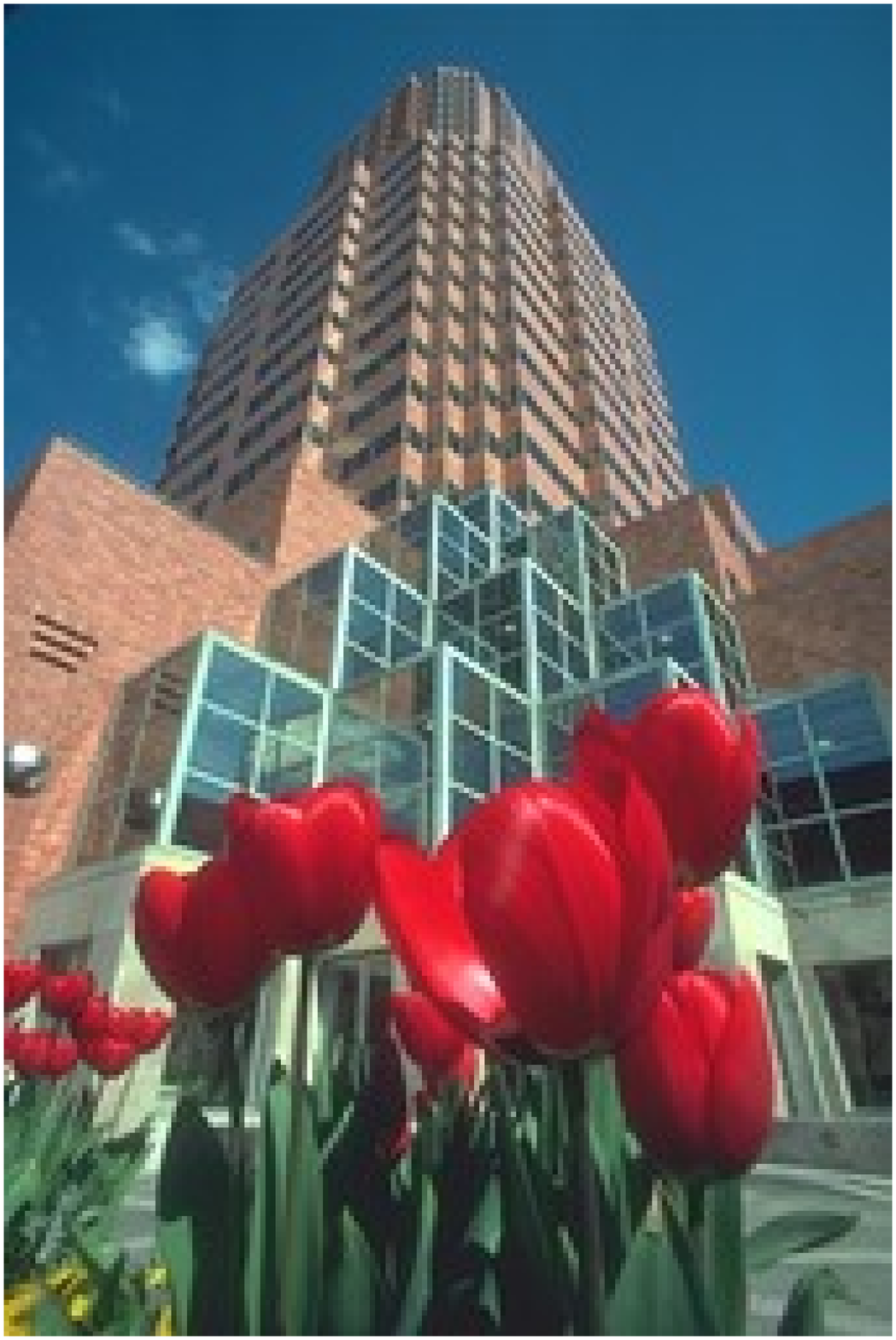}
}
\hfill
\subfigure[$D(R)=S\!E(R)$]{
\includegraphics[width=0.11\linewidth]{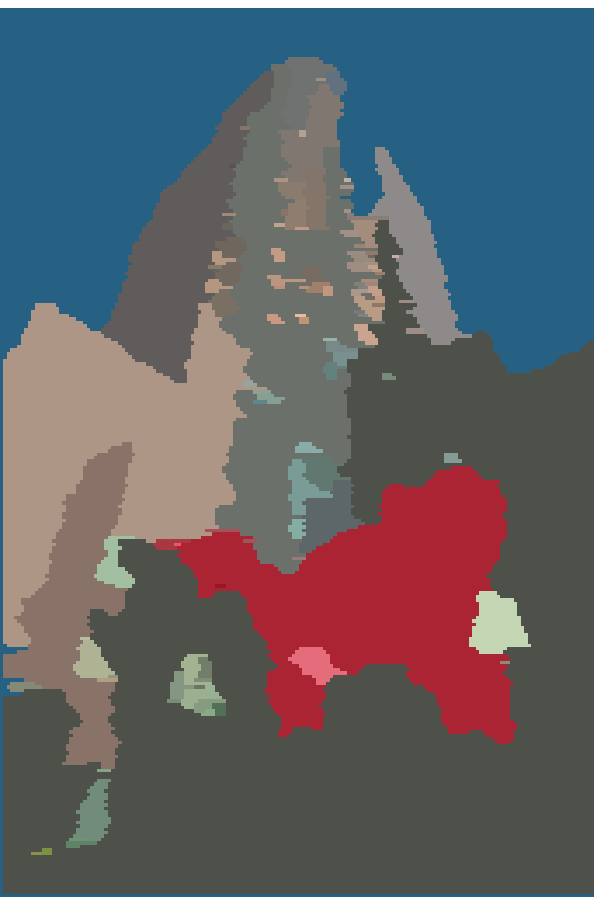}
\includegraphics[width=0.11\linewidth]{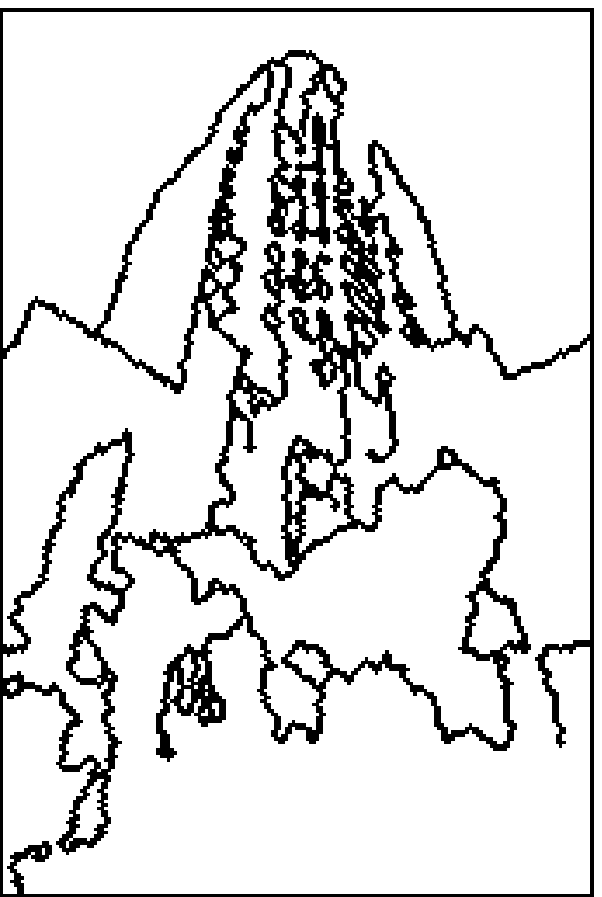}
}
\hfill
\subfigure[$D(R)=SE(R)(1+f(\frac{Int(R)}{Ext(R)})$]{
\includegraphics[width=0.11\linewidth]{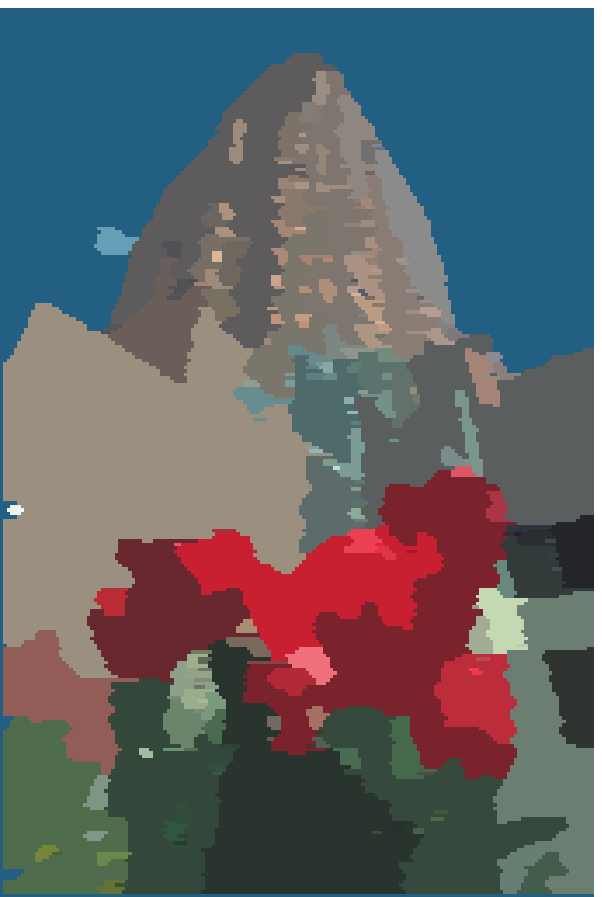} 
\includegraphics[width=0.11\linewidth]{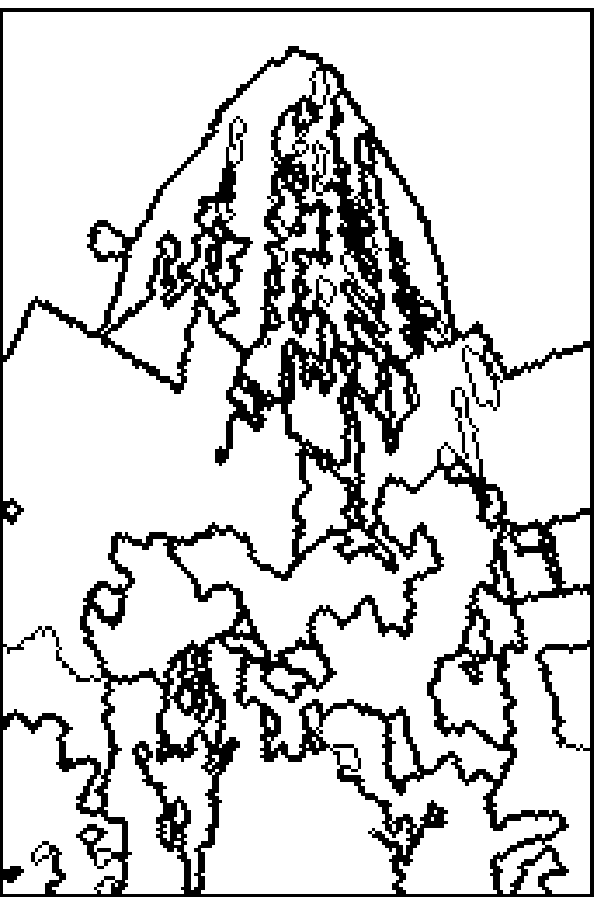}
}
\hfill\mbox{ }
\caption{\label{fig:Contrast} (a) Original images. (b) and (c),
  partitions of the tower image built with a same heuristic(SM) at a
  same normalised scale ($x_\lambda=.8$) but with energies defined using two
  different fit to data terms. (b) is defined using the squared error 
  $D(R)=SE(R)$ while (c) is defined using the formula
  defined by equation~\protect\ref{eq:contrast}.}
\end{figure}

Fig.~\ref{fig:2a2} shows $5$ optimal cuts obtained for increasing
values of $\lambda$ on the Mushroom and Fisherman images of the Berkeley
database\footnote[1]{Color plates are available at the following url:
  http://www.greyc.ensicaen.fr/$\sim$jhpruvot/Cut/} . The heuristics used
to build the hierarchies are displayed on the first column of
Fig.~\ref{fig:2a2}. The original images are displayed in
Fig.~\ref{fig:Contrast}(a).
 
Fig.~\ref{fig:curves}(a) shows the influence of the number of initial 
regions on the execution time. These curves have been obtained on the 
Mushroom image with different initial partitions obtained by varying 
the smoothing parameter of the gradient within our Watershed 
algorithm. 
 
Fig.~\ref{fig:curves}(b) allows to compare the performance of each 
heuristic on the whole Berkeley database. However, a direct comparison 
of the energies obtained by the different heuristics on different 
images would be meaningless since the shape of the function 
$E_\lambda(C^*_\lambda(H))$ depends both of the intrinsic performances of the 
heuristic used to build $H$ and of the image $I$ on which $H$ has been 
built. We have thus to normalise the energies $E_\lambda(C^*_\lambda(H))$ produced 
by the different heuristics before any comparison.

Given a hierarchy $H$, since $C^*_\lambda(H)$ is an unbiased multi-scale 
segmentation (Section~\ref{sec:Guigues}), the hierarchy $H$ obtained 
by each of our methods may be associated to a value $\lambda^{H}_{max}$ 
above which the optimal partition $P_{max}$ is reduced to a single 
region encoding the whole image. The energy of $P_{max}$ is defined 
as: $E_\lambda(P_{max})=D_I+\lambda C_I$ where $D_I=S\!E(I)$ denotes the global 
image's squared error and $C_I=|\delta(I)|$ the perimeter of the image. 
Since the energy of the optimal cuts $E_\lambda(C^*_\lambda(H))$ of a hierarchy 
$H$ is a piecewise linear concave function of $\lambda$, the function 
$E_\lambda(C^*_\lambda(H))$ is below the energy $E_{\lambda}(P_{max})$ associated to the 
coarser partition(Fig.~\ref{fig:curves}(c)).  Moreover, if $P_0$ 
denotes the initial partition, the two points $(0,E_0(P_0))$ and 
$(\lambda_{max},E_{\lambda_{max}}(P_{max}))$ belong to the curve. Therefore, 
$E_\lambda(C^*_\lambda(H))$ being concave, it should be above the line connecting 
these two points. Finally, the line connecting $(0,0)$ to 
$(\lambda_{max},E_{\lambda_{max}}(P_{max}))$ being below the line joining 
$(0,E_0(P_0))$ and $(\lambda_{max},E_{\lambda_{max}}(P_{max}))$ we have for any 
hierarchy $H$ and any scale $\lambda$ (Fig.~\ref{fig:curves}(c)): 
\[
\frac{\lambda}{\lambda_{max}}E_{\lambda_{max}}(P_{max})\leq E_\lambda(C^*_\lambda(H))\leq E_{\lambda}(P_{max}) 
\]
We obtain from this last inequality and after some calculus the
following equation:
\begin{equation} 
  \forall \lambda\in \mathbb{R}+\quad x_\lambda\leq 1+\frac{x_\lambda-1}{1+x_\lambda E_I}\leq \frac{E_\lambda(C^*_\lambda(H))}{E_{\lambda}(P_{max})}\leq 1\mbox{ with } 
x_\lambda=\frac{\lambda}{\lambda_{max}}\mbox{ and }E_I=\frac{\lambda_{max}C_I}{D_I}\label{eq:bounds} 
\end{equation} 
Therefore, using the normalised energy, 
$\frac{E_\lambda(C^*_\lambda(H))}{E_{\lambda}(P_{max})}$ and the normalised scale $x_\lambda 
=\frac{\lambda}{\lambda_{max}}$, any curve $\frac{E_\lambda(C^*_\lambda(H))}{E_{\lambda}(P_{max})}$ 
lies in the upper left part of the unit cube $[0,1]^2$. Note that this 
result is valid for any hierarchy $H$ and thus any heuristic. 
 
Using our piecewise constant model (equation~\ref{eq:mumford}), the 
energy $E_{\lambda}(P_{max})$ is roughly equal to the squared error of the 
image for small values of $\lambda$ and may be interpreted as the global 
variation of the image.  The normalised energy allows thus to reduce 
the influence of the global variation of the images on the energy and 
to compare energies computed with a same heuristic but on different 
images. Note however, that the use of the normalised scale 
$x_\lambda=\frac{\lambda}{\lambda_{max}}$ discards the absolute value of $\lambda_{max}$. We 
thus do not take into account the range of scales for which the 
optimal cut is not reduced to the trivial partition $P_{max}$. 
However, the absolute value of $\lambda_{max}$ varies according to each 
image and each heuristics. The normalised scale allows thus to remove 
the influence of the image. Moreover, our experiments shown thus that 
for each image, our different heuristics obtain close $\lambda_{max}$ 
values. 
 
Fig.~\ref{fig:curves}(b) represents for each value of $x_\lambda$ and each 
heuristic, the mean value of the normalised energy 
$\frac{E_\lambda(C^*_\lambda(H))}{E_{\lambda}(P_{max})}$ computed on the whole set of 
images of the Berckley database. 
 
As shown in Fig~\ref{fig:curves}(b) the energy of the optimal cuts
obtained by the heuristic $MM^1$ (\curveMMUn) is lower than the one
obtained by the maximal matching heuristic (\curveMM). This result is
confirmed by Fig.~\ref{fig:2a2} (lines $MM$ and $MM^1$) where the
heuristic $MM$ removes more details of the mushroom at a given scale.
This result is connected to the greater decimation ratio of the $MM$
heuristic. The $MM$ heuristic merges at each step regions with
important scale of appearance without considering regions which may
appear at further steps.  The algorithms $MM$ and $MM^1$ induce
equivalent execution times on a sequential machine. The execution
times of the method $MM^1$ (\curveMMUn) are overlayed by the ones of
the method $MM$ (\curveMM) in Fig.~\ref{fig:curves}(a) due to the
vertical scale of this figure.

The subjective quality of the partitions obtained by the heuristics
$MM^1$ and $SM^2$ (Fig.~\ref{fig:2a2}) seems roughly similar.  We can
notice that the heuristic $MM^1$ seems to produce slightly coarser
partitions at each scale.  However, considering
Fig.~\ref{fig:curves}(b), the optimal energy obtained by the heuristic
$SM^2$ (\curveSMDeux) are lower than the one obtained by $MM^1$
(\curveMMUn). Note that the heuristic $MM^1$ produces lower execution
times than $SM^2$ even on a sequential
machine(Fig.~\ref{fig:curves}(a)).
 
As shown by Fig.~\ref{fig:curves}(b) the optimal energies produced by
the heuristic $SM$ (\curveSM) are always below the one produced by the
heuristic $SM^2$ (\curveSMDeux).  Note that, the curve (\curveSM) is
close to the diagonal of the square $[0,1]^2$. This last point
indicates that on most of the images of the Berkeley database the
hierarchies produced by the $SM$ heuristic provide optimal cuts whose
normalised energy is closed from the lower bound of the optimal cut's
energies (equation~\ref{eq:bounds}). This result is confirmed by
Fig.~\ref{fig:2a2} where the heuristic $SM$ preserves more details of
the image at each scale.  However, the heuristic $SM$ is the one which
requires the more important execution times on a sequential machine
(Fig.~\ref{fig:curves}(a)).
 
The heuristic $SM^5$ may be understood as a compromise between $SM^2$ 
and $SM$. As shown by Fig.~\ref{fig:curves}(b) the optimal energies 
obtained by the heuristic $SM^5$ (\curveSMCinq) are close to the one 
obtain by $SM$(\curveSM) and below the one obtained by 
$SM^2$(\curveSMDeux). Moreover, as shown by Fig.~\ref{fig:curves}(a), 
the execution times required by $SM^5$ are between the one required by 
the heuristics $SM^2$ and $SM$.  Finally, the partitions obtained by 
the $SM^5$ heuristic in Fig.~\ref{fig:2a2} are closed from the one 
obtained by the heuristic $SM$.\\

Fig.~\ref{fig:Contrast} shows results obtained using an other fit to
data criterion based on the intuitive notion of contrast. The basic
idea of this criterion~\cite{Felzenszwalb} states that a region should
have a higher contrast with its neighbours (called external contrast)
than within its eventual subparts (called internal contrast). Let us
denote by $G_e$ the mean gradient computed along the contour
associated to an edge $e$. The internal and external contrasts of a
region $R$ are then respectively defined as $Int(R)=max_{e\in CC(R)}G_e$
and $Ext(R)=min_{e\in E|v\in \iota(e)}G_e$. Where $CC(R)$ denotes the set of
edges which have been contracted to define $R$ and ${e\in E|v\in \iota(e)}$
denotes the set of edges incident to $v$. Our new energy combines the
contrast and the squared error criteria as follows:
\begin{equation}
  E_\lambda(P)=\sum_{i=1}^nSE(R_i)\left(1+f\left(\frac{Int(R_i)}{Ext(R_i)}\right)\right)+\lambda|\delta(R_i)|\label{eq:contrast}
\end{equation}
where $f()$ denotes a sigmoid function.

A contrasted region will thus have a low ratio between its internal and
external contrast. Conversely, a poorly contrasted region may have a
fit to data term close to twice its squared error.  As shown by
Fig.~\ref{fig:curves}(b) and (c) this energy favours highly contrasted
regions. For example, the cloud merged with the sky in
Fig.~\ref{fig:curves}(b) remains in Fig.~\ref{fig:curves}(c).
Moreover, experiments not reported here, shown us that the same type
of discussion about the advantages and drawbacks of the different
heuristics may be conducted on this new energy with the same
conclusions.


\section{Conclusion}
\label{sec:Conclusion}

The Scale Set framework is based on two steps: the determination of a
hierarchy according to an energy criterion and the determination of
optimal cuts within this hierarchy.  We have presented in this article
parallel and sequential heuristics to build such hierarchies.  The
normalised energy of the optimal cuts, associated with these hierarchy
are bounded bellow by the diagonal of the unit square $[0,1]^2$. Our
experimental results suggest that our sequential heuristic $SM$
provides hierarchies whose normalised energies are closed from this
lower bound. This methods may however require important execution
times. We thus propose an alternative heuristic providing lower
execution time at the price of generally slightly higher optimal cut's
energies. Our parallel methods provide greater energies than the one
produced by Guigues's heuristic. However, these methods require less
execution times even on sequential machine.

Hierarchies encoding a sequence of optimal cuts are usually composed
of a lower number of levels and regions than the initial hierarchies
built by our merge heuristics. In the future, we would like to use
these hierarchies of optimal cuts in order to match two hierarchies
encoding the content of two images sharing a significant part of a
same scene.


\bibliography{gbr2007}

\begin{thebibliography}{10}

\bibitem{lecellier-06}
Lecellier, F., Jehan-Besson, S., Fadili, M., Aubert, G., Revenu, M., Saloux,
  E.:
\newblock Region-based active contours with noise and shape priors.
\newblock In: proceedings of ICIP'2006. (2006)  1649--1652

\bibitem{geman-84}
Geman, S., Geman, D.:
\newblock Stochastic relaxation, gibbs distribution, and the bayesian
  restoration of images.
\newblock IEEE Transactions on PAMI. \textbf{6}(6) (1984)  721--741

\bibitem{leclerc-89}
Leclerc, Y.G.:
\newblock Constructing simple stable descriptions for image partitioning.
\newblock International Journal of Computer Vision \textbf{3}(1) (1989)
  73--102

\bibitem{boykov-04}
Boykov, Y., Kolmogorov, V.:
\newblock An experimental comparison of min-cut/max-flow algorithms for energy
  minimization in vision.
\newblock IEEE Transaction on PAMI \textbf{26}(9) (2004)  1124--1137

\bibitem{guigues03}
Guigues, L., Cocquerez, J.P., Men, H.:
\newblock Scale-sets image analysis.
\newblock Int. J. Comput. Vision \textbf{68}(3) (2006)  289--317

\bibitem{mumfordShah89}
Mumford, D., Shah, J.:
\newblock Optimal approximation by piecewise smooth functions and associated
  variational problems.
\newblock Communications on Pure Applied Mathematics \textbf{42} (1989)
  577--685

\bibitem{haxhimusa-03}
Haxhimusa, Y., Glantz, R., Kropatsch, W.:
\newblock Constructing stochastic pyramids by mides - maximal independent
  directed edge set.
\newblock In Hancock, E., Vento, M., eds.: Proc. of GbR'2003. Volume 2726 of
  LNCS. (2003)  35--46

\bibitem{maxMatching04}
Biedl, T., Demaine, E.D., Duncan, C.A., Fleischer, R., Kobourov, S.G.:
\newblock Tight bounds on maximal and maximum matching.
\newblock Discrete Mathematics \textbf{285}(Issues 1-3) (2004)  7--15

\bibitem{jolion-01}
{J}olion, J.M.:
\newblock Data driven decimation of graphs.
\newblock In Jolion, J.M., Kropatsch, W., Vento, M., eds.: Proceedings of
  $3^{rd}$ IAPR-TC15 Workshop on Graph based Representation in Pattern
  Recognition, Ischia-Italy (2001)  105--114

\bibitem{brun-05}
Brun, L., Mokhtari, M., Meyer, F.:
\newblock Hierarchical watersheds within the combinatorial pyramid framework.
\newblock In: Proc. of DGCI 2005. Volume 3429., IAPR-TC18, LNCS (2005)  34--44

\bibitem{Felzenszwalb}
Felzenszwalb, P., Huttenlocher, D.:
\newblock Image segmentation using local variation.
\newblock In: In Proceedings of IEEE Conference on CVPR, Santa Barbara, CA.
  (1998)  98--104

\end{thebibliography}
\bibliographystyle{splncs}

\end{document}